\documentclass{article}
\usepackage{spconf,amsmath,graphicx,hyperref}

\usepackage{multirow}

\usepackage{amsfonts}
\usepackage{graphicx}
\usepackage{subcaption}

\usepackage[square,sort,comma,numbers]{natbib}
\title{Optimal Transport Based Unsupervised Restoration Learning Exploiting Degradation Sparsity}
\name{Fei Wen$^{1}$, Wei Wang$^{1}$, Zeyu Yan$^{1}$, Wenbin Jiang$^{2}$\thanks{Code is available at \url{https://anonymous.4open.science/r/SOT}.}
}
\address{{$^{1}$School of Information Science and Electronic Engineering, Shanghai Jiao Tong University, China} \\
{$^{2}$School of Communication Engineering, Hangzhou Dianzi University, Hangzhou, China}}
\begin{document}
%\ninept
%
\maketitle
\begin{abstract}
Optimal transport (OT) has recently been shown as a promising criterion for unsupervised restoration when no explicit prior model is available. Despite its theoretical appeal, OT still significantly falls short of supervised methods on challenging tasks such as super-resolution, deraining, and dehazing. In this paper, we propose a \emph{sparsity-aware optimal transport} (SOT) framework to bridge this gap by leveraging a key observation: the degradations in these tasks exhibit distinct sparsity in the frequency domain. Incorporating this sparsity prior into OT can significantly reduce the ambiguity of the inverse mapping for restoration and substantially boost performance. We provide analysis to show exploiting degradation sparsity benefits  unsupervised restoration learning. Extensive experiments on real-world super-resolution, deraining, and dehazing demonstrate that SOT offers notable performance gains over standard OT, while achieving superior perceptual quality compared to existing supervised and unsupervised methods. In particular, SOT consistently outperforms existing unsupervised methods across all three tasks and narrows the performance gap to supervised counterparts. 
\end{abstract}
\section{Introduction}
\label{sec:intro}

Image restoration plays a fundamental 
role in low-level computer vision.
%, which has attracted 
%much research attention in the past few decades.
% Traditional model based methods typically utilize 
% prior models of image and/or degradation, such as
% sparsity, low-rankness, smoothness, self-similarity priors of natural image, and Gaussianity, spatially 
% independent i.i.d. of noise \cite{nlm,bm3d,elad1997restoration,dsc,he2010single}.
Benefited from
deep learning techniques, the field of image restoration has made
much progress in the past a few years \cite{zhang2020residual}.
Generally, the success of deep network based methods usually rely on sufficient
paired degraded-clean data. Nevertheless, in some applications
such paired data is difficult or even impractical
to collect. For such applications, as a common practice, 
synthesized degraded-clean data can be used to learn a restoration model.
However, the gap between synthesized and real-world data
limits the performance on real-world data.
%,
%especially when the degradation is too complex to accurately simulate.
In these circumstances, unsupervised methods without 
requiring any degraded-clean pairs are preferred.

% To relax the requirement of paired training data,
% unsupervised learning methods have recently been
% actively studied for various image restoration tasks.
% Recently,
% unsupervised learning methods have  been
% actively studied.
For unsupervised/self-supervised denoising
learning, the noise-to-noise (N2N) \cite{n2n}, 
noise-to-void (N2V) \cite{n2v}, noise-to-self (N2S) \cite{n2s} methods,
as well as many variants have been developed.
% , see \cite{mbvd,s2s,esur,laine2019high,wu2020unpaired} and the references therein.
While N2N uses noisy pairs, N2V and N2S learn the restoration 
models in a self-supervised manner. Typically, these 
methods rely on prior assumptions, such as the noise in noisy pairs 
are independent \cite{n2n,esur}, or the noise is spatially independent 
and/or the noise type is a priori known \cite{n2v,n2s}.
Such assumptions limit the performance when the noise
does not well conform with the assumptions.

% In the recent work \cite{wang2022optimal},
% it has been shown that, in the absence of any 
% prior model of the degradation,
% the unsupervised restoration learning problem 
% can be optimally formulated as an optimal transport (OT) problem.
More recently,
it has been shown that
the unsupervised restoration learning problem 
can be ideally formulated as an optimal transport (OT) problem \cite{wang2022optimal}.
The OT method has 
achieved promising performance on various denoising 
tasks to approach supervised methods,
but still significantly lags behind 
supervised methods on more complex restoration tasks 
such as super-resolution, deraining, and dehazing.
A main reason is that, for these restoration tasks,
seeking an inverse of the degradation map is more ambiguous.
This work is motivated to
exploit prior models of the degradation to 
learn better inverse maps in the OT framework for restoration.
%reduce the ambiguity in learning 
%an inverse map of the degradation.
We exploit the sparsity 
of degradation in the OT framework to  
boost the performance on these complex restoration tasks. 
%The main contributions are as follows.

%First, we disclose an empirical observation that, for the 
% super-resolution, deraining, and dehazing tasks, the degradation is 
% quite sparse in the frequency domain. Based on this observation,
% we propose a sparsity-aware optimal transport (SOT) criterion for 
% unsupervised restoration learning. 

First, we disclose that, for the 
super-resolution, deraining, and dehazing tasks, the degradation is 
quite sparse in the frequency domain. Then, based on this observation,
we propose a sparsity-aware optimal transport (SOT) criterion for 
unsupervised restoration learning. 
Further, we provide analysis to show the advantage of the proposed SOT criterion.
Finally, we provide extensive experimental results on both synthetic 
and real-world data for the super-resolution, deraining, 
and dehazing tasks. The results demonstrate that 
exploiting the sparsity of degradation can significantly 
boost the performance of the OT criterion.
For example, compared with the vanilla OT criterion 
on real-world super-resolution, deraining and dehazing, 
it achieves a PSNR improvement of about 2.6 dB, 2.7 dB and 1.3 dB, respectively,
and at meantime, attains the best perception scores 
among the compared supervised and unsupervised methods.
Noteworthily, on all the tasks, SOT considerably outperforms existing unsupervised methods to approach the performance of supervised methods.

\section{Preliminaries}

%This section first presents the problem formulation 
%and then introduces the OT based unsupervised 
%restoration learning method.

% \subsection{Problem Formulation}

Consider a degradation to restoration process \(X~\rightarrow ~Y ~\stackrel{f}{\rightarrow} ~\hat{X}\),
where $X \sim {p_X}$ is the source, 
$Y$ is the degraded observation,
$\hat X: = f(Y)$ is the restoration with $f$
being the restoration model to be learned.
Without loss of generality, 
we consider a typical additive model for 
the degradation as \(Y = X + N\),
% \begin{equation}
% Y = X + N,
% \label{eq_degradation}
% \end{equation}
where $N$ stands for the degradation.
Generally, due to the information loss in the data 
process chain $X\rightarrow Y$,
seeking an inverse process from $Y$ to $X$ for 
restoration is ambiguous.

\subsection{OT for Unsupervised Restoration Learning}

% OT can be traced back to the seminal work of 
% Monge \cite{monge1781memoire} in 1781, 
% with significant advancements by Kantorovich 
% \cite{kantorovich1942translation} in 1942. 
% Computationally, the OT problem seeks the most efficient transport map of transforming one distribution of mass to another with minimum cost \cite{Villani2003}.
% Specifically, let ${\cal P}(X)$ and ${\cal P}(Y)$ denote two sets of probability measures on $X$ and $Y$, respectively. 
% Meanwhile, let $\nu  \sim {\cal P}(Y)$ and $\mu \sim {\cal P}(X)$ denote two probability measures.
% The OT problem seeks the most efficient transport plan from $\nu $ to $\mu$ that minimizes the transport cost.

Let ${\cal P}(X)$ and ${\cal P}(Y)$ denote two sets of probability measures on $X$ and $Y$, respectively. 
Meanwhile, let $\nu  \sim {\cal P}(Y)$ and $\mu \sim {\cal P}(X)$ denote two probability measures.
The OT problem seeks the most efficient transport plan from $\nu $ to $\mu$ that minimizes the transport cost.

\textbf{Definition 1. (Transport map)}: 
Given two probability measures $\nu \sim {\cal P}(Y)$ 
and $\mu  \sim {\cal P}(X)$,
$f:Y \to X$ is a transport map from $\nu$ to $\mu$ if
\begin{equation}
\mu (A) = \nu ({f^{ - 1}}(A)) \nonumber,
\end{equation}
for all $\mu$-measurable sets $A$.

\textbf{Definition 2. (Monge’s optimal transport problem) \cite{monge1781memoire}}: 
For a probability measure $\nu$, let ${f_{\sharp}} \nu$ denote the transport of $\nu$ by $f$.
Let $c:Y \times X \to [0, + \infty ]$ be a cost function that $c(y,x)$ measures the cost of transporting $y \in Y$ to $x \in X$. Then, given two probability measures $\nu \sim {\cal P}(Y)$ and $\mu \sim {\cal P}(X)$, the OT problem is defined as%find a $\nu$-measurable map $f:Y \to X$ by
\begin{equation}
%\begin{array}{c}
\mathop {\inf }\limits_f \int_Y {c\left( {f(y),y} \right)} d\nu (y)\quad
{\rm{subject \ to\ }}~~\mu  = {f_\# }\nu ,
%\end{array}
\label{eq_MongeOT}
\end{equation}
over $\nu$-measurable maps $f:Y \to X$.
A minimum to this problem, denoted by ${f^*}$, 
is called an OT map from $\nu$ to $\mu$.

% The OT problem seeks a transport plan to turn the mass of 
% $\nu $ into $\mu $ at the minimal geometric cost measured 
% by the cost function $c$, e.g. typically 
% $c\left( {f(y),y}\right): = {\left\|{f(y)-y} \right\|^\beta}$ with $\beta\ge 1$.

In the absence of any degraded-clean pairs for supervised 
learning of the restoration model, an unsupervised learning
method using only unpaired noisy and clean data has been 
proposed in \cite{wang2022optimal} based on the OT 
formulation (\ref{eq_MongeOT}) as
\begin{equation}
%\begin{gathered}
  \mathop {\min }\limits_f {\mathbb{E}_{Y }}\left(\left\|Y-f(Y)\right\|^\beta\right) \quad
  {\text{subject \ to \ }}~~{p_{\hat X}} = {p_X},
%\end{gathered}
\label{eq_ot}
\end{equation}
with $\beta=2$. 
Formulation (\ref{eq_ot}) is an OT problem seeks a 
restoration map $f$ from $Y$ to $X$. 
The constraint ${p_{\hat X}} = {p_X}$ enforces 
that the restoration $\hat X$ 
has the same distribution as natural clean image $X$.
Under this constraint, the restoration would have good perception quality, 
i.e. looks like natural clean images, since each $\hat X$
lies in the manifold (set) of natural clean image $X$ 
\cite{blau2018perception}. %\cite{blau2018perception,yan2021perceptual,blau2019rethinking, yan2022perceptual}. 
Meanwhile, the constraint ${p_{\hat X}} = {p_X}$
imposes an image prior stronger than any hand-crafted priors,
such as sparsity, low-rankness, smoothness and dark-channel prior.
%These priors have been widely used in traditional 
%image restoration methods. 
Any reasonable prior model
for natural clean images would be fulfilled under 
the constraint ${p_{\hat X}} = {p_X}$.

In implementation, the OT based formulation \eqref{eq_ot} 
is relaxed into an unconstrained form as
\begin{equation}
\mathop {\min }\limits_f {{\mathbb E}_{Y }}\left(\left\|Y-f(Y)\right\|^\beta\right) + \lambda \cdot d({p_X},{p_{\hat X}}),
\label{eq_ot_unconstrained}
\end{equation}
where $\lambda > 0$ is a balance parameter. 
Then, formulation (\ref{eq_ot_unconstrained})
is implemented based on WGAN-gp \cite{wgangp}.
%Though relaxed, it has been shown in 
%\cite{wang2022optimal} that under certain 
%conditions the unconstrained formulation 
%(\ref{eq_ot_unconstrained}) has the same 
%solution as the original constrained 
%formulation (\ref{eq_ot}) in theory.
%However, in practice the balance parameter $\lambda$
%needs to be tuned to achieve satisfactory performance.
% More recently, an OT algorithm for
% unpaired super-resolution has been
% proposed in \cite{ot-sr2022}, which is 
% an alternative for solving (\ref{eq_ot}).

\section{Proposed SOT Method}

% The degradation map $X\rightarrow Y$  is typically non-injective and
% inevitably incurs information loss of the source $X$.
% Hence, it is non-invertible and seeking
% an inverse of it is ambiguous. Exploiting
% prior information of the degradation map is an effective way
% to reduce the inverse ambiguity. 
% In this section, we first
% show the sparsity property of the degradation for three restoration tasks, and then exploit this property
% in the OT framework to propose a sparsity-aware
% formulation for restoration learning. 

%Furthermore, we provide an analytic
%example to show the effectiveness of exploiting sparsity in
%reducing the ambiguity of inverting the degradation process.

\subsection{The Sparsity of Degradation}
\label{degradation_sparsity}

\begin{figure}[t]

\begin{minipage}[b]{0.28\linewidth}
  \centering
  ~~~~~~\centerline{\includegraphics[width=2.8cm]{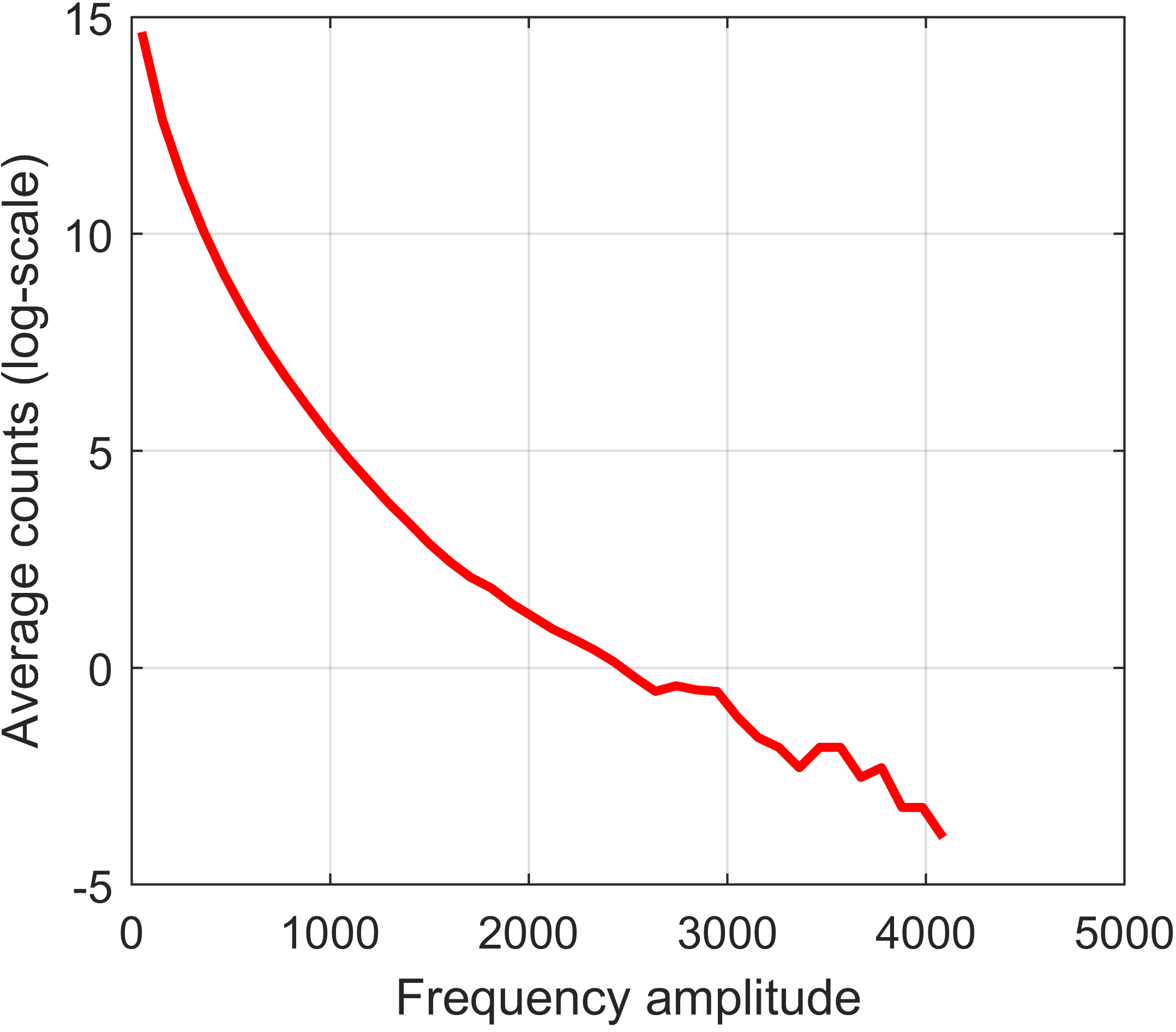}}
%  \vspace{2.0cm}
  \centerline{(a) SR}
\end{minipage}~~~~
\begin{minipage}[b]{.28\linewidth}
  \centering
  \centerline{\includegraphics[width=2.8cm]{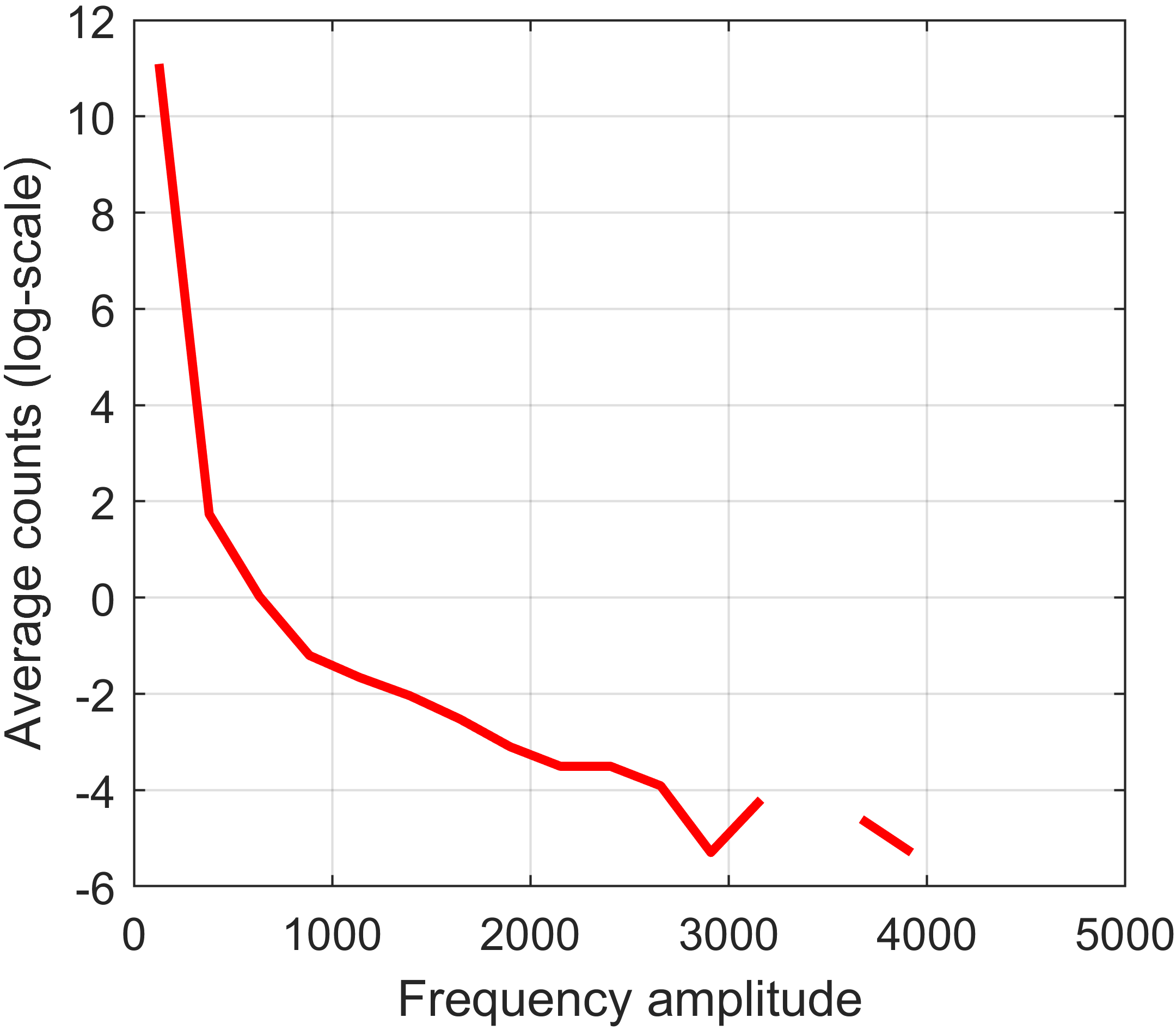}}
%  \vspace{1.5cm}
  \centerline{(b) Deraining}
\end{minipage}~~~~
\begin{minipage}[b]{0.28\linewidth}
  \centering
  \centerline{\includegraphics[width=2.8cm]{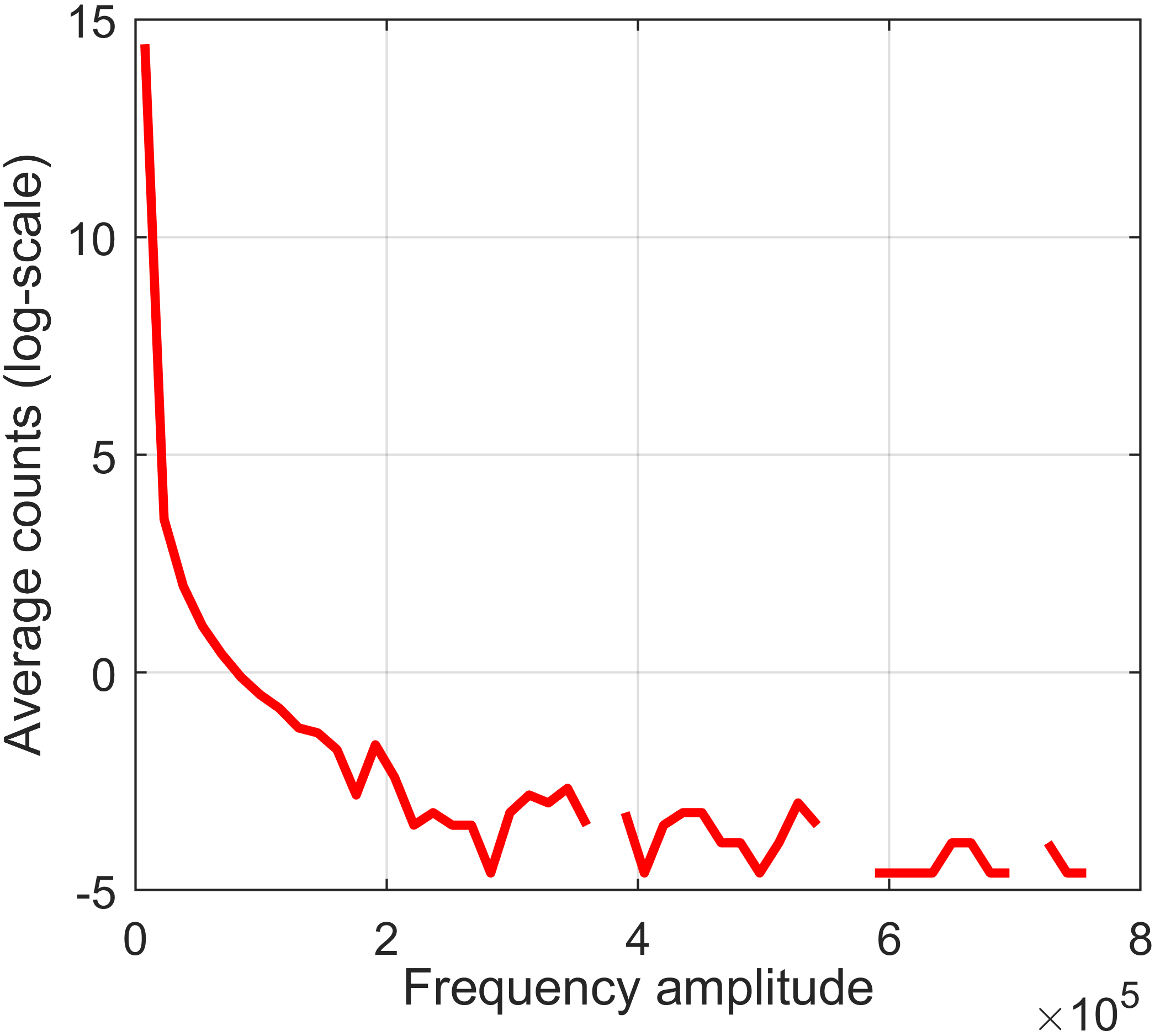}}
%  \vspace{1.5cm}
  \centerline{(c) Dehazing}
\end{minipage}
\vspace{-.2cm}
\caption{Histograms of the degradation $N$ in the frequency domain. (a) Super-resolution (RealVSR \cite{realVSR}). (b) Deraining (SPA \cite{spa}). (c) Dehazing (Dense-haze  \cite{dense}).}
\label{fig_sparse}
\vspace{-10pt}
\end{figure}

Figure \ref{fig_sparse} presents the histogram statistic of the
degradation in the frequency domain for each of the three tasks.
500 real-world images from the super-resolution dataset RealVSR \cite{realVSR},
400 real-world images from the deraining dataset SPA \cite{spa}, and 200
real-world images from the dehazing dataset Dense-haze \cite{dense} are used to
compute the histogram statistic for the three tasks, respectively.
Given $L$ degraded-clean pairs $\{(y_i,x_i)\}_{i=1,\cdots,L}$ from a dataset,
the histogram in the frequency domain is computed based on
the absolute FFT coefficients $\{|{\rm{fft2}}(y_i-x_i)|\}_{i=1,\cdots,L}$,
${\rm{fft2}}(\cdot)$ is the 2D FFT. Each histogram is averaged over the image
pair number $L$. For super-resolution, $Y$ is obtained from a $1/4$ times
low resolution version of $X$ by 4 times bicubic up-sampling.
Figure \ref{fig_sparse} shows that, for each of the three tasks,
the frequency domain representation of the degradation is very sparse.
This significant sparsity feature of the degradation provides
a natural and useful prior for restoration inference.

% It can be seen from Figure \ref{fig_sparse} that, for each of the three tasks,
% the frequency domain representation of the degradation is very sparse,
% which follows a hyper-Laplacian distribution.
% This significant sparsity feature of the degradation provides
% a natural and useful prior for restoration inference.
% It is exploited in our method by promoting sparsity of the
% degradation in the frequency domain to achieve more accurate restoration.
% Note that although these statistic results are derived based on
% real-world degradation data, the implementation of our algorithm
% does not require an accurate estimation of the sparsity parameter
% of the degradation and, as will be shown in experiments,
% a rough selection of $q$ of the $\ell_q$ cost
% (e.g. $q\in\{0.5,1\}$) is enough for achieving satisfactory performance.

\subsection{Exploiting Degradation Sparsity in OT Framework}

The OT criterion (\ref{eq_ot}) does not impose any
constraint on the degradation $N$. 
%Though optimal in the case without prior information of the degradation, 
Therefore, it is suboptimal when
prior information of the degradation is available. 
Based on the above observation, we propose
a formulation to make use of the frequency domain sparsity of degradation as
\begin{equation}
\begin{gathered}
  \mathop {\min }\limits_f {\mathbb{E}_{Y }}\left(\left\|\mathcal{F}(f(Y)-Y)\right\|^q_q\right) ~~
  {\text{subject\ to }}~~{p_{\hat X}} = {p_X},
\end{gathered}
\label{eq_sot}
\end{equation}
where $\mathcal{F}(\cdot)$ stands for the discrete Fourier transform,
and $\|\cdot\|^q_q$ is the $\ell_q$-norm with $0\leq q\leq 1$.
Since the FFT representation of $N$, i.e. $\mathcal{F}(N)$, is sparse,
a necessary condition for an inverting map $f$ to be optimal is that
it should conform to this property such that $\mathcal{F}(N)=\mathcal{F}(f(Y)-Y)$ is sparse.
To achieve this, in formulation (\ref{eq_sot}) we use the $\ell_q^q$ cost with $0\leq q\leq 1$ to promote the sparsity. The $\ell_q$-norm
is a sparsity-promotion loss, which has been widely used
in sparse recovery. % to obtain sparse solution. % \cite{lq}.

For $q=2$, formulation (\ref{eq_sot}) reduces
to the OT problem (\ref{eq_ot}) with $\ell_2$ cost since
$\|\mathcal{F}(f(Y)-Y)\|^2_2=\|f(Y)-Y\|^2_2$.
The $\ell_2^2$ cost is optimal for degradation with Gaussian distribution
but not optimal for sparse degradation with super-Gaussian distribution.
As it has been shown in Figure \ref{fig_sparse} that, the frequency domain distribution
of degradation is far from Gaussian rather being hyper-Laplacian, using
$\ell_2^2$ cost can result in suboptimal performance.
In implementation, an unconstrained form of (\ref{eq_sot}) is used as
\begin{equation}
  \mathop {\min }\limits_f {\mathbb{E}_{Y}}\left(\left\|\mathcal{F}(f(Y)-Y)\right\|^q_q\right) +
  \lambda d(p_{\hat X}, p_X),
\label{eq_sot_unconstrained}
\end{equation}
where $\lambda>0$ is a balance parameter.
As the discrete Fourier transform is complex-valued, for  $M\in\mathbb{C}^{m+1}$, the $\ell_q$ loss is computed as $\|M\|_q^q=\sum_{i=0}^m\left(\Re^2\{M(i)\}+\Im^2\{M(i)\}\right)^{\frac{q}{2}}$.

\subsection{Analysis on the Advantage of SOT}\label{the-anan}

% Our considered OT formulation is not usual as those in the OT community. Specifically, in existing studies on OT, the cost is typically defined on a distance such as the Euclidean distance:
The proposed $\ell_q^q$-cost based OT formulation departs from standard practices in the OT community, where the cost is typically defined on distance metrics, such as the Euclidean distance $c_{Eu}(x,y)\!=\!\rho(\|x\!-\!y\|)$,
% \begin{equation}
% c_{Eu}(x,y)=\rho(\|x-y\|), \label{cost_ot}
% \end{equation}  
where $\|\!\cdot\!\|$ denotes the Euclidean distance. The convex function $\rho(\cdot)=(\cdot)^\beta$ with $\beta\ge 1$ is widely used \cite{Villani2003}, with $\beta=1$ and $\beta=2$  associated with the Wasserstein-1 and Wasserstein-2 distances, respectively. 
Here, we show the advantage of SOT from the model misspecification theory \cite{white1996estimation}.
Let $f_{q}$ denote a solution to the $\ell_q^q$-cost based SOT formulation, and $f_{2}$ denote a solution to the MSE-cost based OT formulation \eqref{eq_ot} with $\beta=2$. Then, we have the following result.
% Meanwhile, denote the restoration of $f_2$ and $f_q$ by
% \[\hat{X}_{ot}:=f_{2}(Y), \quad \text{and} \quad \hat{X}_{sot}:=f_{q}(Y).\] 

\textbf{Theorem 1}. 
Suppose that the degradation $N$ follows a zero-mean GGD $\mathcal{G N}\left(n; 0_{L\times 1}, \sigma \text{I}_{L\times L}, q\right)$ with $q\le 1$. Then, under some regularity conditions, it holds that
\begin{equation}
\mathbb{E}\left[\|f_{q}(Y)-Y\|_q^q\right]
\le
\mathbb{E}\left[\|f_{2}(Y)-Y\|_q^q\right].
\label{lower_risk_lq}
\end{equation}
Meanwhile, denote the restoration residual (estimated noise) of $f_2$ and $f_q$ by
\(\hat{N}_{{ot}}=Y-f_{2}(Y)\), and \(\hat{N}_{{sot}}=Y-f_{q}(Y)\), respectively,
it holds that
\begin{equation}
KL\left(p_N\,|\,p_{\hat{N}_{sot}}\right) \le KL\left(p_N\,|\,p_{\hat{N}_{ot}}\right).
\end{equation}

% \textbf{Theorem 1}. 
% Denote the MSE distortion of the MSE-cost and $\ell_q$-cost based OT restoration by \(D_{ot} = \mathbb{E}\{\|\hat{X}_{ot} - X\|_2^2\}\) and \(D_{sot} = \mathbb{E}\{\|\hat{X}_{sot} - X\|_2^2\}\), respectively.
% Suppose that the degradation $N$ follows a zero-mean GGD $\mathcal{G N}\left(n; 0_{L\times 1}, \sigma \text{I}_{L\times L}, q\right)$ with $q\le 1$. Then, under some regularity conditions, it holds that
% \begin{equation}
% D_{sot} \le D_{ot}.
% \label{lower_distortion}
% \end{equation}

The proof is provided in Appendix. This result indicates that, under the same feasible set defined by the distribution alignment constraint, using the \(\ell_q^q\)-cost (the correct likelihood model under the GGD assumption) leads to a more accurate noise estimation. Consequently, this finding implies that the SOT solution, which uses \(\ell_q^q\) cost, can achieve better restoration than the quadratic cost based OT.

\section{Experimental Results}\label{exp_res}

We conduct evaluation on super-resolution, deraining, and dehazing. For each task, 
SOT is compared with representative supervised and 
unsupervised methods on both synthetic and real-world data.
% Note that, since for each of the tasks there exists a number of
% supervised and unsupervised methods, it is difficult to compare with 
% all the representative methods in each task.
% The focus here is to 
% compare with state-of-the-art supervised and unsupervised methods
% in each task. Particularly, state-of-the-art supervised
% methods are used as ideal baselines for comparison.
For our method, two variants, denoted by SOT ($\ell_{0.5}$) and SOT ($\ell_1$), 
are evaluated in each task, which use the $\ell_{0.5}$ cost and $\ell_{1}$ cost, respectively.
The compared methods are:
\textit{1)} Super-resolution:
RankSR \cite{ranksrgan}, RCAN \cite{rcan}, 
ESRGAN \cite{wang2018esrgan} and RNAN \cite{rnan}. 
The compared unsupervised methods include USIS \cite{usis}, 
OT \cite{wang2022optimal}.
\textit{2)} Deraining: DSC\cite{dsc}, RESCAN\cite{rescan}, 
MPRNet\cite{multi}, SIRR\cite{wei2019semi}, 
CycleGAN\cite{cyclegan}, DeCyGAN\cite{ deraincyclegan}, 
OT\cite{wang2022optimal}, where DSC is a traditional method, 
RESCAN and MPRNet are supervised methods, SIRR is a semi-supervised method, 
while CycleGAN, DeCycleGAN and OT are unsupervised methods.
\textit{3)} Dehazing: DCP \cite{he2010single}, 
AODNet \cite{aodnet}, Dehamer \cite{dehamer}, GCANet \cite{gcanet}, 
FFANet \cite{ffanet}, D4 \cite{ d4}, OT \cite{wang2022optimal}, 
where DCP is a traditional model-based method, 
AODNet, Dehamer, GCANet and FFANet are supervised learning methods, 
wile D4 and OT are unsupervised learning methods. 

% In order to make a comprehensive evaluation, the restoration quality
% is evaluated in terms of both distortion metrics, 
% including peak signal to  noise  ratio  (PSNR) and structural similarity (SSIM),
% and perceptual quality metrics, including perception index (PI) \cite{pi} and 
% learned perceptual image patch similarity (LPIPS) \cite{lpips}.

The performance 
is evaluated in terms of PSNR, SSIM, PI  and 
LPIPS.
We implement (\ref{eq_sot_unconstrained}) based on 
WGAN-gp \cite{wgangp}. Note that the best selection of $q$ is application and data dependent.
In the implementation we only roughly test two values of $q$, 
e.g., $q=0.5$ and $q=1$ for the $\ell_q^q$ cost of SOT.
Experimental results show that this rough selection is 
sufficient to yield satisfactory performance of SOT.
%Although accurate estimation of the sparsity parameter of the residuals can effectively improve the reconstruction, these statistic results are derived based on real-world degradation data. So in the implementation, only the parameters $q=0.5$ and $q=1$ are roughly chosen to construct the $\ell_q$ loss, and the experiments verify that this approach is sufficient to achieve satisfactory reconstruction results.
For a fair comparison, our method and the OT method \cite{wang2022optimal}
use the same network structure, which consists of a generator and a discriminator. 
The generator uses the network in MPRNet \cite{multi},
%, which is one of the state-of-the-art models, 
while the discriminator is the same as that in \cite{wang2022optimal}. 
% All the experiments are performed on a PC with a
% single RTX-3090 GPU with 24GB memory. 

\begin{table}[!t]
    \scriptsize
    \centering
    \caption{Comparison on 4x super-resolution for both synthetic images (DIV2K) and real-world images (RealVSR).}
    \begin{tabular}{|c|cc|c|c|}
        \hline
        \textbf{Case} & \multicolumn{2}{c|}{\textbf{Method}} & \textbf{PSNR/SSIM} & \textbf{LPIPS/PI} \\ \hline
        
        \multirow{9}{*}{Synthetic} 
        & \textbf{Traditional} & Bicubic & 26.77/0.755 & 0.1674/7.07 \\ \cline{2-5}
        & \multirow{4}{*}{\textbf{Supervised}} & RankSR  & 26.56/0.734 & 0.0541/\textbf{3.01} \\ \cline{3-5}
        &  & RCAN    & \textbf{29.33}/\textbf{0.828} & 0.0925/5.24 \\ \cline{3-5}
        &  & ESRGAN  & 26.66/0.749 & \textbf{0.0520}/3.26 \\ \cline{3-5}
        &  & RNAN    & 29.31/0.827 & 0.0966/5.40 \\ \cline{2-5}
        & \multirow{4}{*}{\textbf{Unsupervised}} & USIS    & 22.22/0.628 & 0.1761/3.52 \\ \cline{3-5}
        &  & OT     & 25.73/0.719 & 0.0838/4.44 \\ \cline{3-5}
        &  & SOT ($\ell_{0.5}$)   & 28.25/0.801 & 0.0612/3.03 \\ \cline{3-5}
        &  & SOT ($\ell_1$)  & 27.91/0.783 & 0.0821/3.45 \\ \hline

        \multirow{9}{*}{Real-world} 
        & \textbf{Traditional} & Bicubic & 23.15/0.749 & 0.1347/5.94 \\ \cline{2-5}
        & \multirow{4}{*}{\textbf{Supervised}} & RankSR  & 21.05/0.607 & 0.1031/\textbf{2.79} \\ \cline{3-5}
        &  & RCAN    & 23.39/0.772 & 0.1573/5.79 \\ \cline{3-5}
        &  & ESRGAN  & 21.13/0.632 & 0.1024/3.53 \\ \cline{3-5}
        &  & RNAN    & 23.19/0.752 & 0.1599/5.88 \\ \cline{2-5}
        & \multirow{4}{*}{\textbf{Unsupervised}} & USIS    & 19.06/0.502 & 0.2122/3.35 \\ \cline{3-5}
        &  & OT     & 21.34/0.658 & 0.1110/4.12 \\ \cline{3-5}
        &  & SOT ($\ell_{0.5}$)   & \textbf{23.89}/\textbf{0.782} & 0.0839/3.27 \\ \cline{3-5}
        &  & SOT ($\ell_1$)  & 22.85/0.722 & \textbf{0.0758}/3.23 \\ \hline
    \end{tabular}
    \label{table:1}
    \vspace{-10pt}
\end{table}

\textbf{Image Super-Resolution.}
We conduct super-resolution experiment on both synthetic dataset DIV2K\cite{div2k} and real-world dataset RealVSR \cite{realVSR}.
Table \ref{table:1} presents results of 4x super-resolution. On synthetic data,  SOT
can achieve PSNR and SSIM results close to those of 
supervised learning methods, e.g., with about 1.06 dB difference 
compared with RCAN. Meanwhile, in terms of the perception indices,
its perceptual quality exceeds some of the supervised methods 
such as RCAN. 
% RankSR and ESRGAN 
% use perceptual quality metrics such as LPIPS as the optimization 
% target, hence can achieve 
% better perceptual quality scores but worse PSNR and SSIM results. 
% However, recent studies on the trade-off between distortion and 
% perceptual quality show that, the improvement in perceptual quality 
% would necessarily lead to increase of reconstruction distortion, 
% e.g. the deterioration of the MSE, PSNR and SSIM metrics 
% \cite{blau2018perception}. Accordingly, 
% the PSNR and SSIM results of RankSR and ESRGAN are worse. 
Moreover, SOT significantly outperforms the
vanilla OT method, e.g. an improvement of about 2.52 dB in PSNR. 
On the real data, SOT can achieve better 
PSNR, SSIM and LPIPS scores even compared with 
supervised methods. SOT($\ell_{0.5}$) achieves a PSNR about 2.55 dB higher than OT
with significant better perception scores.
%reconstruction quality since SOT gets higher scores than OT. 
SOT($\ell_{0.5}$) has lower distortion than SOT($\ell_{1}$) 
(e.g. about 1 dB higher in PSNR) but with worse perception scores.
This  accords well with the distortion-perception tradeoff theory.

\textbf{Image Deraining.}
For synthetic image deraining, we train the models on 
the Rain1800 dataset \cite{zhang2019image} and test on 
the Rain100L dataset \cite{yang2017deep}.
From Table \ref{table:2}, SOT can achieve a PSNR close to the 
supervised method. For example, the difference in PSNR between SOT($\ell_{1}$) 
and MPRNet is about 1.07 dB. Noteworthily, SOT($\ell_{1}$) 
achieves the best perception scores among all the compared unsupervised 
and supervised methods.
Again, SOT performs much better than OT, which demonstrates the 
effectiveness sparsity exploitation. 
% For SOT, the $\ell_{1}$ cost
% yields better performance than the $\ell_{0.5}$ cost.
For real-world image derainging, 
we use the dataset SPA \cite{spa}. 
Similar to the synthetic experiment,
SOT achieves the best performance among the unsupervised methods. 
It achieves a PSNR only 1.75 dB lower than  
the supervised method MPRNet.
Particularly, SOT achieves the best perception scores, 
which even surpasses that of the supervised methods.
In addition, SOT($\ell_1$) attains a PSNR 2.69 dB higher than
that of the vanilla OT method.

\begin{table}[!t]
    \scriptsize
    \centering
    \caption{Comparison on deraining  for both synthetic (Rain1800 and Rain100L) and real-world data (SPA).}
    \begin{tabular}{|c|cc|c|c|}
        \hline
        \textbf{Case} & \multicolumn{2}{c|}{\textbf{Method}} & \textbf{PSNR/SSIM} & \textbf{LPIPS/PI} \\ \hline
        
        \multirow{11}{*}{Synthetic} 
        & \textbf{} & Rainy & 25.52/0.825 & 0.1088/3.77 \\ \cline{2-5}
        & \textbf{Traditional} & DSC & 25.72/0.831 & 0.1116/2.79 \\ \cline{2-5}
        & \multirow{2}{*}{\textbf{Supervised}} & RESCAN & 29.80/0.881 & 0.0731/2.87 \\ \cline{3-5}
        &  & MPRNet & \textbf{36.40}/\textbf{0.965} & 0.0167/3.21 \\ \cline{2-5}
        & \textbf{Semi-supervised} & SIRR & 23.48/0.800 & 0.0978/2.88 \\ \cline{2-5}
        & \multirow{5}{*}{\textbf{Unsupervised}} & CycleGAN & 24.03/0.820 & 0.0960/2.80 \\ \cline{3-5}
        &  & DeCyGAN & 24.89/0.821 & 0.0952/3.12 \\ \cline{3-5}
        &  & OT & 33.71/0.954 & 0.0158/2.76 \\ \cline{3-5}
        &  & SOT ($\ell_{0.5}$) & 34.74/0.948 & 0.0132/2.54 \\ \cline{3-5}
        &  & SOT ($\ell_{1}$) & 35.33/0.963 & \textbf{0.0108}/\textbf{2.51} \\ \hline
        
        \multirow{11}{*}{Real-world} 
        & \textbf{} & Rainy & 34.30/0.923 & 0.0473/8.49 \\ \cline{2-5}
        & \textbf{Traditional} & DSC & 32.29/0.921 & 0.0498/7.89 \\ \cline{2-5}
        & \multirow{2}{*}{\textbf{Supervised}} & RESCAN & 38.34/0.961 & 0.0250/8.03 \\ \cline{3-5}
        &  & MPRNet & \textbf{46.12}/\textbf{0.986} & 0.0109/7.68 \\ \cline{2-5}
        & \textbf{Semi-supervised} & SIRR & 22.66/0.710 & 0.1323/7.87 \\ \cline{2-5}
        & \multirow{5}{*}{\textbf{Unsupervised}} & CycleGAN & 28.79/0.923 & 0.0422/7.58 \\ \cline{3-5}
        &  & DeCyGAN & 34.78/0.929 & 0.0528/7.50 \\ \cline{3-5}
        &  & OT & 41.68/0.951 & 0.0098/7.35 \\ \cline{3-5}
        &  & SOT ($\ell_{0.5}$) & 42.69/0.958 & 0.0096/7.15 \\ \cline{3-5}
        &  & SOT ($\ell_{1}$) & 44.37/0.982 & \textbf{0.0084}/\textbf{7.06} \\ \hline
    \end{tabular}
    \label{table:2}
    \vspace{-10pt}
\end{table}

\textbf{Dehazing.}
For synthetic and real-world image dehazing, we use the OTS dataset \cite{li2018benchmarking} and Dense-haze dataset \cite{dense}, respectively. 
% This dataset contains a large number of images of outdoor 
% scenes with various levels of synthetic fog layers. We 
% selected 100 images from the dataset as the test set.
Table \ref{table:3} shows the results of the compared 
methods on the two datasets. On synthetic data, SOT can achieve a PSNR approaching that of the supervised methods.
For instance, the difference in PSNR between SOT($\ell_{1}$) 
and the transformer based supervised method Dehamer is about 1.18 dB, while SOT($\ell_{1}$) achieves 
the best perception scores among all the compared 
supervised and unsupervised methods.
In this experiment, SOT with the $\ell_1$ cost attains a 
PSNR 3.37 dB higher than the standard OT method, e.g.,
32.20 dB versus 28.83 dB. 

% For the real-world image dahazing task, 
% we chose the real scene dataset Dense-haze \cite{dense} for training and testing. 
% This dataset was obtained from two sets of images in the same scene 
% with fog and under normal conditions through artificial smoke. 
The artificial smoke in Dense-haze is quite dense and hence the 
restoration task is extremely challenging.
Similar to the synthetic dehazing task, SOT achieves the 
best performance among the unsupervised methods. Besides, in term of the
LPIPS and PI scores, its perception quality even surpasses that of the 
supervised methods, i.e. the best among all the compared supervised and 
unsupervised methods. Compared with the standard OT method, the performance 
of SOT still improves considerably on this challenging realistic task.

%The visual quality of different methods is illustrated in Appendix \ref{SynDehsamples}.
% Fig. \ref{figure6} compares the visual quality of the methods. 
% The restoration quality of the proposed method is even  
% on par with Dehamer.

\begin{table}[!t]
    \scriptsize
    \centering
    \caption{Comparison on dehazing for both synthetic data (OTS) and real-world data (Dense-Haze).}
    \begin{tabular}{|c|cc|c|c|}
        \hline
        \textbf{Case} & \multicolumn{2}{c|}{\textbf{Method}} & \textbf{PSNR/SSIM} & \textbf{LPIPS/PI} \\ \hline

        \multirow{10}{*}{Synthetic} 
        & \textbf{} & Hazy & 18.13/0.851 & 0.0747/2.96 \\ \cline{2-5}
        & \textbf{Traditional} & DCP & 16.83/0.863 & 0.0670/2.27 \\ \cline{2-5}
        & \multirow{4}{*}{\textbf{Supervised}} & AODNet & 18.42/0.828 & 0.0703/2.69 \\ \cline{3-5}
        &  & Dehamer & \textbf{33.38}/\textbf{0.946} & 0.0168/2.35 \\ \cline{3-5}
        &  & GCANet & 19.85/0.704 & 0.0689/2.39 \\ \cline{3-5}
        &  & FFANet & 30.80/0.935 & 0.0182/2.68 \\ \cline{2-5}
        & \multirow{4}{*}{\textbf{Unsupervised}} & D4 & 21.89/0.845 & 0.0466/2.39 \\ \cline{3-5}
        &  & OT & 28.83/0.919 & 0.0236/2.60 \\ \cline{3-5}
        &  & SOT ($\ell_{0.5}$) & 31.63/0.905 & 0.0165/2.14 \\ \cline{3-5}
        &  & SOT ($\ell_{1}$) & 32.20/0.935 & \textbf{0.0157}/\textbf{2.08} \\ \hline

        \multirow{10}{*}{Real-world} 
        & \textbf{} & Hazy & 10.55/0.435 & 0.2057/7.04 \\ \cline{2-5}
        & \textbf{Traditional} & DCP & 11.01/0.415 & 0.3441/5.91 \\ \cline{2-5}
        & \multirow{4}{*}{\textbf{Supervised}} & AODNet & 10.64/0.469 & 0.2453/6.11 \\ \cline{3-5}
        &  & Dehamer & \textbf{16.63}/\textbf{0.585} & 0.1523/5.65 \\ \cline{3-5}
        &  & GCANet & 12.46/0.454 & 0.2524/5.04 \\ \cline{3-5}
        &  & FFANet & 8.77/0.452 & 0.1985/6.71 \\ \cline{2-5}
        & \multirow{4}{*}{\textbf{Unsupervised}} & D4 & 9.69/0.462 & 0.2140/5.03 \\ \cline{3-5}
        &  & OT & 14.17/0.503 & 0.1699/4.89 \\ \cline{3-5}
        &  & SOT ($\ell_{0.5}$) & 15.01/0.526 & 0.1523/4.56 \\ \cline{3-5}
        &  & SOT ($\ell_{1}$) & 15.40/0.567 & \textbf{0.1451}/\textbf{4.50} \\ \hline
    \end{tabular}
    \label{table:3}
\end{table}

More detailed experimental settings,  more experiment results and qualitative comparison are provided in Appendix.

% more experiment results and qualitative  comparison are provided in  supplemental material,
% including the effect of the parameter $\lambda$ and visual comparison of different methods.

% Fig. \ref{figure7} compares the visual quality of the methods. 
% Noteworthily, the visual quality of SOT is distinctly better 
% compared with the state-of-the-art transformer based supervised method Dehamer,
% e.g. the color of the palette restored by SOT is much closer to the real scene.
% This task is quite challenging as the observation (hazy images) 
% is severely degraded with an average PSNR of only about 10 dB.
% The results demonstrate the potential of the proposed method
% on realistic difficult tasks to handle complex degradation.

\section{Conclusions}\label{conclusion}

An unsupervised  learning method for image restoration
has been developed, which exploits the sparsity of
degradation in the OT framework to 
reduce the ambiguity in restoration
Extensive evaluation on super-resolution, deraining, and dehazing tasks demonstrates that, it can significantly
improve the performance of the OT criterion to approach 
the performance of supervised methods.
Particularly, among the compared supervised and unsupervised methods, 
the proposed method achieves the best 
PSNR, SSIM and LPIPS results in real-world super-resolution,
and the best perception scores (LPIPS and PI) in 
real-world deraining and dehazing.
Our method is the first generic unsupervised method
that can achieve favorable performance in comparison with
representative supervised methods on all the super-resolution, 
deraining, and dehazing tasks.

\bibliographystyle{IEEEbib}
\bibliography{IEEEabrv, sot}

% \newpage

\appendix

\section{An Analytical Example on the Effectiveness of the Sparsity Prior}
\label{sec4example}

Under the degradation process $Y=X+N$, the distribution of $Y$
is a convolution of the distributions of $X$ and $N$,
i.e., $p_Y=p_X\otimes p_N$. Generally, the inverse problem
from $Y$ to $X$ is ambiguous due to the information loss
in the degradation process. In the absence of any prior
information of $N$, the optimal transport map $f_{\sharp}p_Y=p_X$ obtained
by OT can be used as an ideal inverse (restoration) map \cite{wang2022optimal}.
However, when prior information of $N$ is available, the map $f$ is
no longer optimal and does not necessarily approach the lower bound
of inverse distortion. Naturally, the prior information of $N$ can be
exploited to reduce the inverting ambiguity to some extent and hence
result in lower restoration distortion.

As our objective is to exploit the underlying sparsity of $N$ to reduce
the inverse ambiguity, here we provide an analytical example to show the effectiveness
of exploiting the sparsity of the degradation in helping to find the
desired correct inverse map.

\textbf{Example 1 [Exploiting the sparsity of degradation helps to
find an inverse map with lower distortion].}
Consider a discrete random source $X\in \mathbb{R}^{m+1}$,
which follows a two-point distribution with probability mass function as
\begin{equation}\notag
p_X(x) = \left\{\! {\begin{array}{*{20}{l}}
\!{p_1,}&{x=x_1}\\
\!{p_2,}&{x=x_2}
\end{array}} \right.,
\end{equation}
with
\[x_1=[-a,\mathop{\underbrace{b,\cdots,b}}\limits_{m}]^T,~~~~x_2=[-a,\mathop{\underbrace{-b,\cdots,-b}}\limits_{m}]^T,\]
where $a\gg b>0$ such that $a^2>mb^2$ and $a^q<mb^q$ for any $0\leq q\leq 1$.
For example, these two conditions hold for $a=1$, $b=0.1$ and $10<m<100$.
If only considering $q=0$, they hold for $a=1$, $b=0.1$ and $1<m<100$.
Furthermore, consider a degradation process as \(Y=X+N\),
where $N\in \mathbb{R}^{m+1}$ also follows a two-point distribution
with probability mass function given by
\begin{equation}\notag
p_N(n) = \left\{\! {\begin{array}{*{20}{l}}
\!{\tilde{p}_1,}&{n=n_1}\\
\!{\tilde{p}_2,}&{n=n_2}
\end{array}} \right.,
\end{equation}
with
\[n_1=[-2a,0,\cdots,0]^T,~~~~n_2=[2a,0,\cdots,0]^T.\]
Note that $N$ is sparse as only one of its elements is nonzero.
In this setting, the distribution of $Y$ is given by the convolution
between $p_X$ and $p_N$ as
\begin{equation}\notag
p_Y(y) = \left\{\! {\begin{array}{*{20}{l}}
\!{p_1\tilde{p}_1,}&{y=y_1}\\
\!{p_1\tilde{p}_2,}&{y=y_2}\\
\!{p_2\tilde{p}_1,}&{y=y_3}\\
\!{p_2\tilde{p}_2,}&{y=y_4}
\end{array}} \right.,
\end{equation}
with
\[y_1=[-3a,b,\cdots,b]^T,~~~~y_2=[a,b,\cdots,b]^T,\]
\[y_3=[-a,-b,\cdots,-b]^T,~~~~y_4=[3a,-b,\cdots,-b]^T.\]

Then, given the distributions of $X$ and $Y$, we investigate
the inverse process $Y\rightarrow X$ by seeking the OT
from $p_Y$ to $p_X$ with different cost functions.
Particularly, we evaluate the $\ell_q^q$ cost with $0\leq q \leq 1$
in comparison with the widely used $\ell_2^2$ cost.

\begin{figure*}[!t]  
	\centering  
	% 子图1  
	\begin{subfigure}[b]{0.2\linewidth}  
		\centering  
		\includegraphics[width=\linewidth]{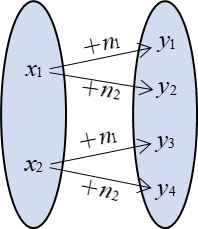}  
		\caption{Degradation process}  
	\end{subfigure}  
	\quad\quad\quad%\hfill % 添加间距  
	% 子图2  
	\begin{subfigure}[b]{0.2\linewidth}  
		\centering  
		\includegraphics[width=\linewidth]{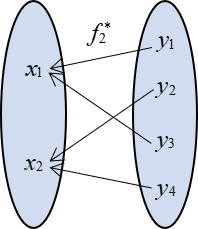}  
		\caption{Inverse via OT with $\ell_2$ cost}  
	\end{subfigure}  
	\quad\quad\quad%\hfill % 添加间距  
	% 子图3  
	\begin{subfigure}[b]{0.2\linewidth}  
		\centering  
		\includegraphics[width=\linewidth]{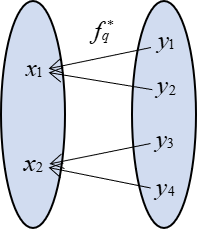}  
		\caption{Inverse via OT with $\ell_q$ cost for any $0\leq q \leq 1$}  
	\end{subfigure}  
	\caption{An illustration of the optimal transport from $p_Y$ to  
$p_X$ with different cost function in Example 1. (a) The degradation process $Y=X+N$. (b) Inverse via the optimal transport $f_2$ with $\ell_2^2$ cost. (c) Inverse via the optimal transport $f_q$ with $\ell_q^q$ cost for any $0\leq q \leq 1$. In this example, the optimal transport map under the $\ell_q^q$ cost yields the correct inverse map with zero distortion, while that under the $\ell_2^2$ cost does not.}  
	\label{fig_example}  
\end{figure*}

First, with the $\ell_2$ cost and under the condition $a^2>mb^2$,
a solution $f_2^*:Y\rightarrow X$ to the OT problem
\begin{equation}
\begin{gathered}
  \mathop {\min }\limits_{f_2} {\mathbb{E}_{Y }}\left(\left\|f_2(Y)-Y\right\|^2_2\right) \\
  {\text{subject \ to \ }}~~(f_2)_{\sharp}{p_Y} = {p_X},
\end{gathered}
\label{eq_sot_examp_f2}
\end{equation}
would map $y_2\rightarrow x_2$ rather than $y_2\rightarrow x_1$ since
$\|y_2-x_2\|^2_2=4mb^2<\|y_3-x_2\|^2_2=4a^2$ and $y_3\rightarrow x_1$
rather than $y_3\rightarrow x_2$ since $\|y_3-x_1\|^2_2=4mb^2<\|y_3-x_2\|^2_2=4a^2$.
For example, let $p_1=\tilde{p}_1=0.5$ and $p_2=\tilde{p}_2=0.5$,
the optimal map $f_2^*$ is given by $f_2^*(y_1)=x_1$, $f_2^*(y_2)=x_2$,
$f_2^*(y_3)=x_1$, and $f_2^*(y_4)=x_2$, as illustrated in Fig. \ref{fig_example}(b).
In contrast, with the $\ell_{q}^q$ cost and under the condition $a^q<mb^q$, $\forall q\in[0,1]$,
it follows that $\|y_2-x_2\|^q_q=m(2b)^q>\|y_3-x_2\|^q_q=(2a)^q$.
Hence, the solution $f_q^*:Y\rightarrow X$ to the OT problem
\begin{equation}
\begin{gathered}
  \mathop {\min }\limits_{f_q} {\mathbb{E}_{Y }}\left(\left\|f_q(Y)-Y\right\|^q_q\right) \\
  {\text{subject \ to \ }}~~(f_q)_{\sharp}{p_Y} = {p_X},
\end{gathered}
\label{eq_sot_examp_fq}
\end{equation}
is given by $f_q^*(y_1)=x_1$, $f_q^*(y_2)=x_1$ and
$f_q^*(y_3)=x_2$, $f_q^*(y_4)=x_2$, 
as illustrated in Figure \ref{fig_example}(c).

% \begin{figure}[!t]
% 	\centering
% 	\subfigure[Degradation process]
%          {
% 		\includegraphics[width=0.28\linewidth]{./Figures/fig_example_a.png}
% 	  }~~~~
%          \subfigure[Inverse via OT with $\ell_2$ cost]
%          {
% 		\includegraphics[width=0.28\linewidth]{./Figures/fig_example_b.png}
% 	}~~~~
%          \subfigure[Inverse via OT with $\ell_q$ cost for any $0\leq q \leq 1$]{
% 		\includegraphics[width=0.28\linewidth]{./Figures/fig_example_c.png}
% 	}
% 	\caption{An illustration of the optimal transport from $p_Y$ to
% $p_X$ with different cost function in Example 1. (a) The degradation process $Y=X+N$. (b) Inverse via the optimal transport $f_2$ with $\ell_2$ cost. (c) Inverse via the optimal transport $f_q$ with $\ell_q$ cost for any $0\leq q \leq 1$. In this example, the optimal transport map under the $\ell_q$ cost yields the correct inverse map with zero distortion, while that under the $\ell_2$ cost does not.}
% 	\label{fig_example}
% \end{figure}

In this example, the residual $N$ is very sparse as it has only
one nonzero element. This sparsity property can be exploited by
using a sparsity-promotion data fitting cost, such as the $\ell_q$-norm
with $0\leq q \leq 1$ \cite{lq}. Since using the $\ell_q^q$ cost
can well exploit the underlying sparsity prior of the degradation
to reduce the ambiguity in finding the inverse map, in this example
it yields the desired correct inverse map under the OT criterion,
which has a zero distortion. In comparison, the $\ell_2$ cost yields
an undesired map, which does not align with the degradation map and
has a nonzero distortion, as shown in Figure \ref{fig_example}.  This example illustrates
that exploiting the sparsity prior of the degradation (e.g., by the
$\ell_q^q$ cost) can effectively help to find an inverse transport map with lower restoration distortion.

\section{Proof of Theorem 1}\label{proof_th2}
Theorem 1 is derived based on the model misspecification theory and the optimal transport theory.
From the condition that the degradation \(N\in\mathbb{R}^L\) follows a hyper‐Laplacian distribution characterized by a zero-mean GGD, $\mathcal{G N}\left(n; 0_{L\times 1}, \sigma \text{I}_{L\times L}, q\right)$ with $q\le 1$, we have
\begin{equation}
p_{N}\left(n\right) \propto  \exp\left(-\frac{\|n\|_q^q}{\sigma}\right), \quad 0 < q \le 1. 
\label{ggdq}
\end{equation}
In the optimal transport formulations, there is a distribution alignment constraint.
With the degradation model \(Y = X + N\), denote the distribution of \(Y\) by \(p_Y\). Then, consider a function space
\begin{equation}
\mathcal{F}
:=\left\{
  f : \mathbb{R}^L \to \mathbb{R}^L
  \;\big|\;
  f_{\sharp}p_Y = p_X
\right\},
\end{equation}
which contains all measurable mappings pushing forward \(p_Y\) to \(p_X\). Denote $\hat{X}=f(Y)$, we have ${p_{\hat X}} = {p_X}$ for any $f\in \mathcal{F}$. With these definitions, the $\ell_q^q$-cost based optimal transport formulation  can be expressed as
\begin{equation}
f_{q} = \arg\underset{f\in \mathcal{F}} {\min } \;{\mathbb{E}_{Y}}\left(\left\|f(Y)-Y\right\|^q_q\right).
\label{eq_sot3}
\end{equation}
Similarly, the MSE-cost based optimal transport formulation (with $\beta=2$) can be expressed as
\begin{equation}
f_{2} = \arg\underset{f\in \mathcal{F}} {\min } \;{\mathbb{E}_{Y}}\left(\left\|f(Y)-Y\right\|^2\right).
\label{eq_sot3}
\end{equation}

Then, the main result of Theorem 1 is derived based on the  model mis-specification theory 
\cite{white1982maximum,white1996estimation}. 
This theory establishes that, under a same feasible set, a model uses the correct likelihood (corresponding to the true noise distribution) has smaller risk (e.g., MSE) than using a mismatched likelihood.

Before proceeding to the main result, we first introduce 
the regularity conditions the model mis-specification theory relies on, which are given in the following assumptions:

\textit{Assumption 1:} 
The feasible set  \(\mathcal{F}:=\{ f | f_{\sharp}p_Y = p_X\}\)
is nonempty, and in \(\mathcal{F}\), the expectations
\({\mathbb{E}_{Y}}(\left\|f(Y)-Y\right\|^q_q)\)
and
\({\mathbb{E}_{Y}}(\left\|f(Y)-Y\right\|^2)\)
each admit at least one global minimizer.

This is a mild assumption.
The existence of an optimal map to the OT formulation has been well studied \cite{gangbo1996geometry,Villani2003}.
Existing results have established the existence of an unique solution to the optimal transport under both convex and concave cost functions \cite{gangbo1996geometry, santambrogio2015optimal,  champion2011monge}.

Define the restoration residual (estimated noise) of $f_{q}$ and $f_{2}$ as 
\[\hat{N}_{\mathrm{sot}} := Y - f_{q}(Y)\] 
and 
\[\hat{N}_{\mathrm{ot}} := Y - f_{2}(Y).\] 
% Then, using
% \[
% \|\hat{X}-X\|_{2}^{2}
% =\|[Y-f(Y)] - N\|_{2}^{2}
% =\|\widehat{N} - N\|_{2}^{2},
% \]
% the restoration distortion of $f_{q}$ and $f_{2}$ can be expressed as
% \begin{equation}
% D_{{sot}}
% =\mathbb{E} \left[\|\widehat{N}_{{sot}} - N\|_{2}^{2}\right],
% \quad
% D_{{ot}}
% =\mathbb{E} \left[\|\widehat{N}_{{ot}} - N\|_{2}^{2}\right].\label{distortion_residual}
% \end{equation}

Under the assumption that the degradation $N$ follows a GGD \eqref{ggdq},
the true negative log‐likelihood is (up to a constant factor) 
\(\|n\|_q^q\). 
Thus, minimizing 
\(\mathbb{E}\left(\|f(Y) - Y\|_q^q\right)\)
under \(f_\sharp p_Y = p_X\) is equivalent to maximizing the likelihood based on the true noise model.
In contrast, \(\|n\|^2\) is a mis-specified model corresponding to a mis-assumed negative log-likelihood under Gaussian distributon.

Denote 
\[\mathcal{L}^q (\cdot)=\|\cdot\|^q_q,\quad \text{and} \quad \mathcal{L}^2 (\cdot)=\|\cdot\|^2.\]
Under the GGD assumption of $N$, \(\mathbb {E}\{\|N\|_q^q\}<\infty\), and the $\ell_q^q$ function \(\|n\|_{q}^q\) satisfies the measurability and integrability. 
Then, with these properties and Assumption 1, and with the fact that $\mathcal{L}^q$ is the correct negative log-likelihood for $N\sim \mathcal{G N}\left(n; 0_{L\times 1}, \sigma \text{I}_{L\times L}, q\right)$ with $q\le 1$, and $\mathcal{L}^2 \neq \mathcal{L}^q$ in this case, it follows from the classical theory of model misspecification \cite{white1982maximum,white1996estimation} that
\begin{equation}
\mathbb{E}\left[\mathcal{L}^q\left(\hat{N}_{{ot}}\right)\right]
\geq
\mathbb{E}\left[\mathcal{L}^q\left(\hat{N}_{{sot}}\right)\right],\label{lower_risk}
\end{equation}

This is a direct result from White’s theorems on mis-specified likelihoods \cite{white1982maximum,white1996estimation}: the estimator that matches the true density’s log-likelihood always gives a lower (or at worst equal) true risk than the one derived from an incorrect model, provided both lie in the same feasible space.
Specifically, within the same feasible space \(f\in \mathcal{F}\) (i.e. with the same constraint \(f_\sharp p_Y = p_X\)),  $\mathcal{L}^2$ using the wrong noise model (the MSE-cost does not match the correct likelihood for GGD noise with $q\le 1$) leads to is suboptimal relative to the true model $\mathcal{L}^q$, which matches the true log-likelihood of \(N\sim \mathcal{G N}\left(n; 0_{L\times 1}, \sigma \text{I}_{L\times L}, q\right)\).

From the definition of $\mathcal{L}^q$, $\hat{N}_{ot}$ and $\hat{N}_{sot}$, \eqref{lower_risk} can be rewritten as
\begin{equation}
\mathbb{E}\left[\|\hat{N}_{{ot}}\|_q^q\right]
\geq
\mathbb{E}\left[\|\hat{N}_{{sot}}\|_q^q\right].
\label{lower_risk_lq}
\end{equation}
Furthermore, since minimizing expected negative log-likelihood is equivalent (up to a constant) to minimizing $KL\left(p_N\,|\,p_{\hat{N}}\right)$, it follows from \eqref{lower_risk} that
\begin{equation}
KL\left(p_N\,|\,p_{\hat{N}_{sot}}\right) \le KL\left(p_N\,|\,p_{\hat{N}_{ot}}\right).
\end{equation}

\section{Experimental Setting and Results}

\subsection{More Detailed Experimental Setting}
We conduct evaluation on super-resolution, deraining, and dehazing. For each task, 
the proposed method is compared with representative supervised and 
unsupervised methods on both synthetic and real-world data.
% Note that, since for each of the tasks there exists a number of
% supervised and unsupervised methods, it is difficult to compare with 
% all the representative methods in each task.
% The focus here is to 
% compare with state-of-the-art supervised and unsupervised methods
% in each task. Particularly, state-of-the-art supervised
% methods are used as ideal baselines for comparison.
For our method, two variants, denoted by SOT ($\ell_{0.5}$) and SOT ($\ell_1$), 
are evaluated in each task, which use the $\ell_{0.5}$ cost and $\ell_{1}$ cost, respectively.
The compared methods are as follows.\vspace{-6pt}
\begin{itemize}%[leftmargin=1em]
\item
For super-resolution, the compared supervised methods 
include RankSR \cite{ranksrgan}, RCAN \cite{rcan}, 
ESRGAN \cite{wang2018esrgan} and RNAN \cite{rnan}. 
The compared unsupervised methods include USIS \cite{usis}, 
OT \cite{wang2022optimal}.%, SOT ($\ell_{0.5}$) and SOT ($\ell_1$), 
%where SOT ($\ell_{0.5}$) and SOT ($\ell_1$) are two variants of 
%our method use the $\ell_{0.5}$ and $\ell_{1}$ cost, respectively.
\vspace{-3pt}
\item 
For deraining, the compared methods 
include DSC\cite{dsc}, RESCAN\cite{rescan}, 
MPRNet\cite{multi}, SIRR\cite{wei2019semi}, 
CycleGAN\cite{cyclegan}, DeCyGAN\cite{ deraincyclegan}, 
OT\cite{wang2022optimal}, where DSC is a traditional method, 
RESCAN and MPRNet are supervised methods, SIRR is a semi-supervised method, 
while CycleGAN, DeCycleGAN and OT are unsupervised methods.
\vspace{-3pt}
\item
For dehazing, the compared methods include DCP \cite{he2010single}, 
AODNet \cite{aodnet}, Dehamer \cite{dehamer}, GCANet \cite{gcanet}, 
FFANet \cite{ffanet}, D4 \cite{ d4}, OT \cite{wang2022optimal}, 
where DCP is a traditional model-based method, 
AODNet, Dehamer, GCANet and FFANet are supervised learning methods, 
wile D4 and OT are unsupervised learning methods. 
\end{itemize}
\vspace{-3pt}
% In order to make a comprehensive evaluation, the restoration quality
% is evaluated in terms of both distortion metrics, 
% including peak signal to  noise  ratio  (PSNR) and structural similarity (SSIM),
% and perceptual quality metrics, including perception index (PI) \cite{pi} and 
% learned perceptual image patch similarity (LPIPS) \cite{lpips}.

The restoration quality
is evaluated in terms of both distortion metrics, including PSNR and SSIM,
and perceptual quality metrics, including perception index (PI) \cite{pi} and 
learned perceptual image patch similarity (LPIPS) \cite{lpips}.
We implement the proposed formulation (\ref{eq_sot_unconstrained}) based on 
WGAN-gp \cite{wgangp}
%, with $0 \leq q \leq 1$ and 
with $\lambda $ being tuned for 
each task. Note that the best selection of the value of $q$ depends on 
the statistics of the data and hence is application and data dependent.
In practice it is generally difficult to select the optimal value of $q$.
Therefore, in the implementation we only roughly test two values of $q$, 
e.g., $q=0.5$ and $q=1$ for the $\ell_q^q$ cost of SOT.
Experimental results show that this rough selection is 
sufficient to yield satisfactory performance of SOT.
%Although accurate estimation of the sparsity parameter of the residuals can effectively improve the reconstruction, these statistic results are derived based on real-world degradation data. So in the implementation, only the parameters $q=0.5$ and $q=1$ are roughly chosen to construct the $\ell_q$ loss, and the experiments verify that this approach is sufficient to achieve satisfactory reconstruction results.
For a fair comparison, our method and the OT method \cite{wang2022optimal}
use the same network structure, which consists of a generator and a discriminator. 
The generator uses the network in MPRNet \cite{multi},
%, which is one of the state-of-the-art models, 
while the discriminator is the same as that in \cite{wang2022optimal}. 
All the experiments are performed on a PC with a
single RTX-3090 GPU with 24GB memory.

For the synthetic super-resolution experiment, we use the DIV2K\cite{div2k} 
dataset, which contains a total of 1000 high-quality RGB 
images with a resolution of about 2K. 100 images are 
used for testing. Since OT requires the input to have 
the same size as the output, we follow the pre-upsampling 
method \cite{wang2020deep} to upsample the low-resolution 
images before feeding them into the network by bicubic.

In synthetic super-resolution experiments, bicubic down-sampling 
is widely used to construct paired training data. However, 
real-word degradation can substantially deviate from bicubic down-sampling.
This limits the performance of the synthetic data learned model on real-world data.
To further verify the performance of the proposed method on real scenes, 
we conduct experiment on a real-world super-resolution dataset RealVSR \cite{realVSR}. 
In this dataset, paired data is constructed by firstly using the 
multi-camera system of iPhone 11 Pro Max to capture images 
of different resolutions in the same scene separately, 
and then adopting post-processing such as color correction and pixel alignment. 
The results on this dataset are shown in Table \ref{table2}.

For synthetic image deraining, we train the models on 
the Rain1800 dataset \cite{zhang2019image} and test on 
the Rain100L dataset \cite{yang2017deep}. These two datasets 
respectively contain 1800 and 100 images of natural scenes 
with simulated raindrops.

For real-world image derainging, 
we chose the real scene dataset SPA \cite{spa} for training and testing. 
This dataset takes images with and without rain in the same scene by 
fixing the camera position, hence the rain-free images can be used as 
the ground-truth for supervised model training.

For the synthetic image dehazing task, we train and test
the models on the OTS dataset \cite{li2018benchmarking}. 
This dataset contains a large number of images of outdoor 
scenes with various levels of synthetic fog layers. We 
selected 100 images from the dataset as the test set.

For the real-world image dahazing task, 
we chose the real scene dataset Dense-haze \cite{dense} for training and testing. 
This dataset was obtained from two sets of images in the same scene 
with fog and under normal conditions through artificial smoke. 
The artificial smoke in the dataset is quite dense and hence the 
restoration task is extremely challenging.

\subsection{Effect of the parameter $\lambda$}

\begin{figure}[h]
	\centering
	\includegraphics[width=0.99\linewidth] {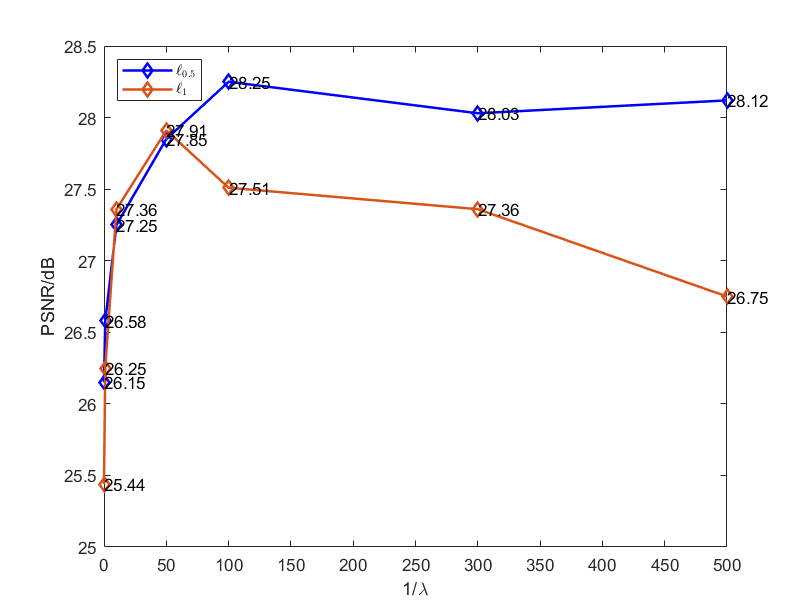}
	\caption{Restoration PSNR of SOT in image super-resolution for different values of $\lambda$.}
	\label{figure2}
\end{figure}

This section provides more experimental results.
Figure \ref{figure2} presents the PSNR of SOT versus the value of $\lambda$ in the  super-resolution experiment,
which shows the effect of the parameter $\lambda$. 

\subsection{Visual Samples of Synthstic Image Super-Resolution}
\label{SynSRsamples}
\begin{figure*}[!t]  
    \centering  
    % 第一行子图  
    \begin{subfigure}[b]{0.3\linewidth}  
        \centering  
        \includegraphics[width=\linewidth]{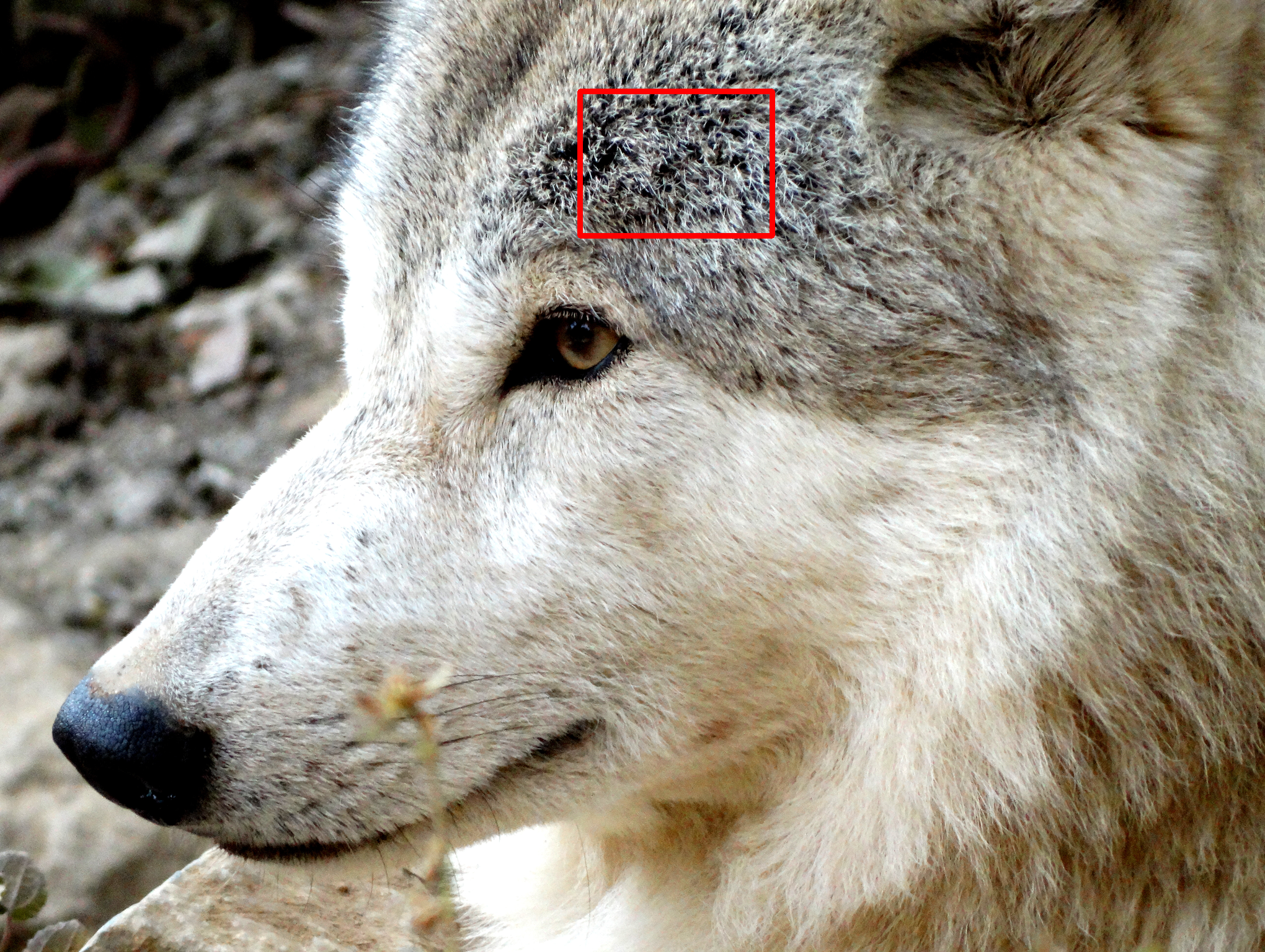}  
        \caption{Test image}  
    \end{subfigure}  
    \hfill  
    \begin{subfigure}[b]{0.3\linewidth}  
        \centering  
        \includegraphics[width=\linewidth]{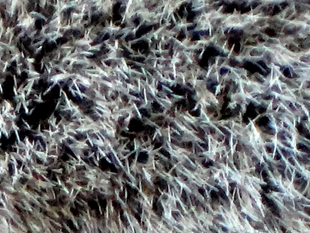}  
        \caption{Ground truth}  
    \end{subfigure}  
    \hfill  
    \begin{subfigure}[b]{0.3\linewidth}  
        \centering  
        \includegraphics[width=\linewidth]{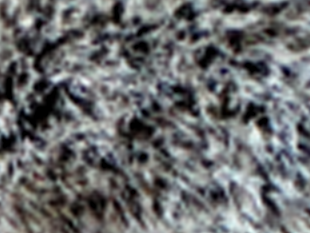}  
        \caption{Bicubic (26.59 dB)}  
    \end{subfigure}  

    % 第二行子图  
    \vspace{0.5em} % 行间距  
    \begin{subfigure}[b]{0.3\linewidth}  
        \centering  
        \includegraphics[width=\linewidth]{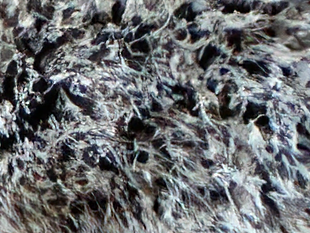}  
        \caption{RankSR (26.64/0.743/0.053)}  
    \end{subfigure}  
    \hfill  
    \begin{subfigure}[b]{0.3\linewidth}  
        \centering  
        \includegraphics[width=\linewidth]{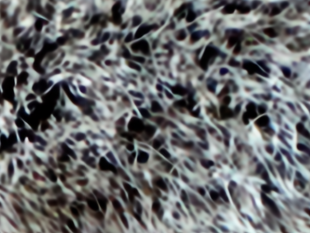}  
        \caption{RCAN (\textbf{29.35}/\textbf{0.829}/0.093)}  
    \end{subfigure}  
    \hfill  
    \begin{subfigure}[b]{0.3\linewidth}  
        \centering  
        \includegraphics[width=\linewidth]{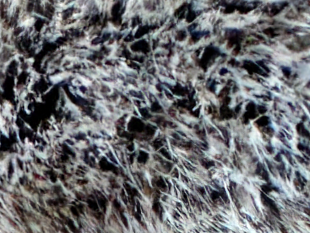}  
        \caption{ESRGAN (26.66/0.751/\textbf{0.052})}  
    \end{subfigure}  

    % 第三行子图  
    \vspace{0.5em} % 行间距  
    \begin{subfigure}[b]{0.3\linewidth}  
        \centering  
        \includegraphics[width=\linewidth]{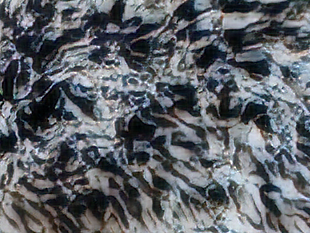}  
        \caption{USIS (23.44/0.701/0.167)}  
    \end{subfigure}  
    \hfill  
    \begin{subfigure}[b]{0.3\linewidth}  
        \centering  
        \includegraphics[width=\linewidth]{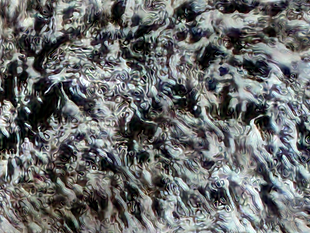}  
        \caption{OT (26.88/0.754/0.082)}  
    \end{subfigure}  
    \hfill  
    \begin{subfigure}[b]{0.3\linewidth}  
        \centering  
        \includegraphics[width=\linewidth]{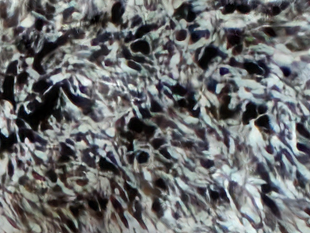}  
        \caption{SOT ($\ell_{0.5}$) (28.62/0.821/0.060)}  
    \end{subfigure}  

    % 总标题  
    \caption{Visual comparison on 4x synthetic image super-resolution. The PSNR/SSIM/LPIPS results are provided in the brackets. The images are enlarged for clarity.}  
    \label{figure1}  
\end{figure*}
Fig. \ref{figure1} compares the visual quality of 4x 
super-resolution on a typical sample from the DIV2K dataset. 
It can be seen that the result of RCAN, which yields the
highest PSNR, is closer to the ground-truth, but the 
reconstructed images appear to be blurred and the 
details are not clear enough. 
The proposed SOT method with the $\ell_{0.5}$ cost yields higher PSNR
than RankSR and ESRGAN, while having comparable perception quality.

\subsection{Visual Samples of Real Image Super-Resolution}
\label{realSRsamples}
\begin{figure*}[!t]  
    \centering  
    % 第一行子图  
    \begin{subfigure}[b]{0.32\linewidth}  
        \centering  
        \includegraphics[width=\linewidth]{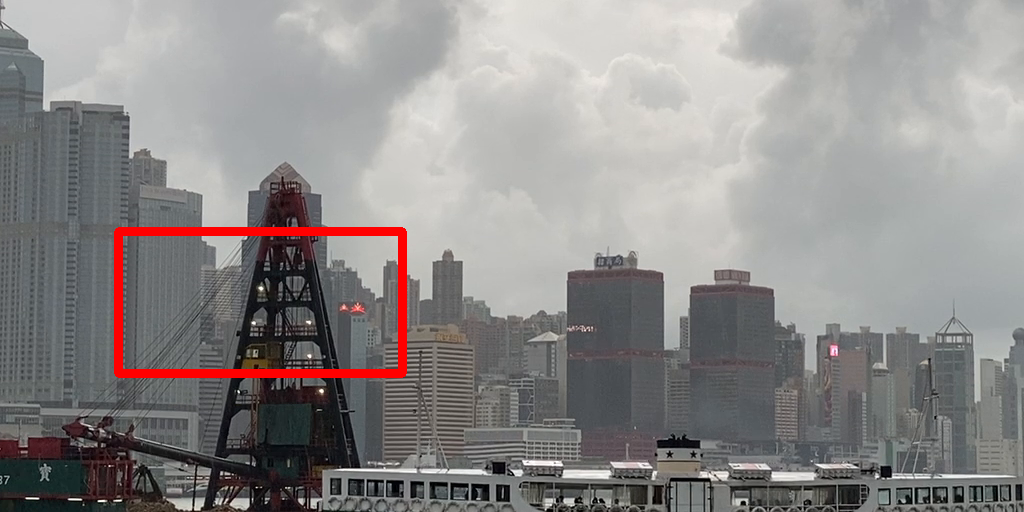}  
        \caption{Test image}  
    \end{subfigure}  
    \hfill  
    \begin{subfigure}[b]{0.32\linewidth}  
        \centering  
        \includegraphics[width=\linewidth]{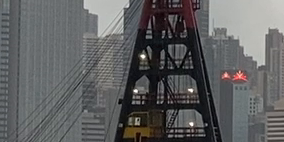}  
        \caption{Ground truth}  
    \end{subfigure}  
    \hfill  
    \begin{subfigure}[b]{0.32\linewidth}  
        \centering  
        \includegraphics[width=\linewidth]{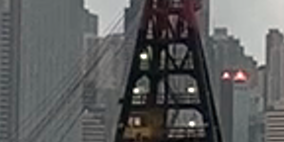}  
        \caption{Bicubic (23.01 dB)}  
    \end{subfigure}  

    % 第二行子图  
    \vspace{0.5em} % 行间距  
    \begin{subfigure}[b]{0.32\linewidth}  
        \centering  
        \includegraphics[width=\linewidth]{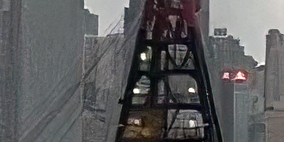}  
        \caption{RankSR (20.87/0.612/0.104)}  
    \end{subfigure}  
    \hfill  
    \begin{subfigure}[b]{0.32\linewidth}  
        \centering  
        \includegraphics[width=\linewidth]{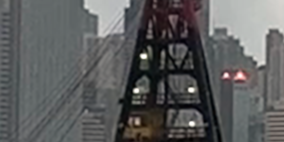}  
        \caption{RCAN (23.48/0.784/0.157)}  
    \end{subfigure}  
    \hfill  
    \begin{subfigure}[b]{0.32\linewidth}  
        \centering  
        \includegraphics[width=\linewidth]{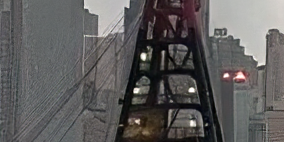}  
        \caption{ESRGAN (21.06/0.634/0.101)}  
    \end{subfigure}  

    % 第三行子图  
    \vspace{0.5em} % 行间距  
    \begin{subfigure}[b]{0.32\linewidth}  
        \centering  
        \includegraphics[width=\linewidth]{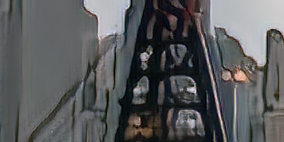}  
        \caption{USIS (19.57/0.527/0.211)}  
    \end{subfigure}  
    \hfill  
    \begin{subfigure}[b]{0.32\linewidth}  
        \centering  
        \includegraphics[width=\linewidth]{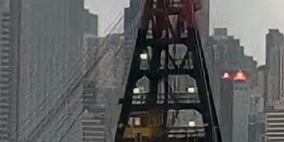}  
        \caption{OT (21.65/0.674/0.111)}  
    \end{subfigure}  
    \hfill  
    \begin{subfigure}[b]{0.32\linewidth}  
        \centering  
        \includegraphics[width=\linewidth]{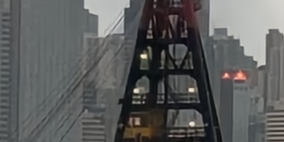}  
        \caption{SOT ($\ell_{0.5}$) (\textbf{24.03}/\textbf{0.792}/\textbf{0.084})}  
    \end{subfigure}  

    % 总标题  
    \caption{Visual comparison on real-world image super-resolution. The PSNR/SSIM/LPIPS results are provided in the brackets. The images are enlarged for clarity.}  
    \label{figure3}  
\end{figure*} 
Fig. \ref{figure3} compares the visual quality of the 
methods on a typical sample from the RealVSR dataset. 
It can be seen that SOT achieves the best 
PSNR, SSIM and LPIPS scores. Qualitatively, 
it can achieve high-quality detail reconstruction
while having less artifacts than the perception-oriented methods.

\subsection{Visual Samples of Synthetic Image Deraining}
\label{SynDersamples}
\begin{figure*}[!t]  
    \centering  
    % 第一行子图  
    \begin{subfigure}[b]{0.3\linewidth}  
        \centering  
        \includegraphics[width=\linewidth]{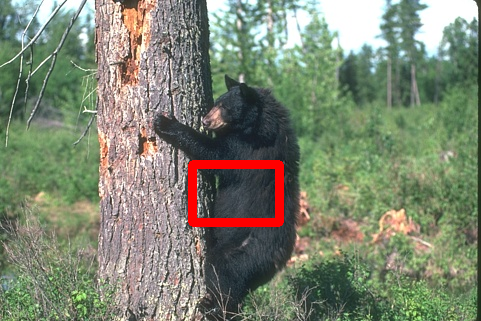}  
        \caption{Test image}  
    \end{subfigure}  
    \hfill  
    \begin{subfigure}[b]{0.3\linewidth}  
        \centering  
        \includegraphics[width=\linewidth]{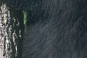}  
        \caption{Ground truth}  
    \end{subfigure}  
    \hfill  
    \begin{subfigure}[b]{0.3\linewidth}  
        \centering  
        \includegraphics[width=\linewidth]{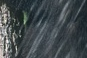}  
        \caption{Rainy (25.41 dB)}  
    \end{subfigure}  

    % 第二行子图  
    \vspace{0.5em} % 行间距  
    \begin{subfigure}[b]{0.3\linewidth}  
        \centering  
        \includegraphics[width=\linewidth]{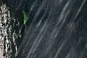}  
        \caption{DSC (25.78/0.835/0.112)}  
    \end{subfigure}  
    \hfill  
    \begin{subfigure}[b]{0.3\linewidth}  
        \centering  
        \includegraphics[width=\linewidth]{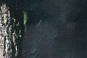}  
        \caption{RESCAN (30.21/0.892/0.072)}  
    \end{subfigure}  
    \hfill  
    \begin{subfigure}[b]{0.3\linewidth}  
        \centering  
        \includegraphics[width=\linewidth]{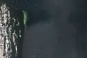}  
        \caption{MPRNet (\textbf{36.58}/\textbf{0.972}/0.018)}  
    \end{subfigure}  

    % 第三行子图  
    \vspace{0.5em} % 行间距  
    \begin{subfigure}[b]{0.3\linewidth}  
        \centering  
        \includegraphics[width=\linewidth]{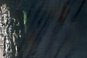}  
        \caption{SIRR (24.59/0.825/0.098)}  
    \end{subfigure}  
    \hfill  
    \begin{subfigure}[b]{0.3\linewidth}  
        \centering  
        \includegraphics[width=\linewidth]{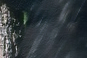}  
        \caption{DeCyGAN (25.32/0.865/0.095)}  
    \end{subfigure}  
    \hfill  
    \begin{subfigure}[b]{0.3\linewidth}  
        \centering  
        \includegraphics[width=\linewidth]{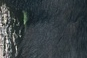}  
        \caption{SOT ($\ell_{1}$) (36.02/0.968/\textbf{0.011})}  
    \end{subfigure}  

    % 总标题  
    \caption{Visual comparison on synthetic image deraining. The PSNR/SSIM/LPIPS results are provided in the brackets. The images are enlarged for clarity.}  
    \label{figure4}  
\end{figure*}
Fig. \ref{figure4} compares the visual quality of the methods.
It can be observed that the supervised methods, such as MPRNet, 
can achieve excellent rain removal but with the restoration being over-smoothing.
The proposed method can reconstruct better texture details, 
while achieving effective rain removal.

\subsection{Visual Samples of Real-world Image Deraining}
\label{realDersamples}
\begin{figure*}[!t]  
    \centering  
    % 第一行子图  
    \begin{subfigure}[b]{0.3\linewidth}  
        \centering  
        \includegraphics[width=\linewidth,height=0.7\linewidth]{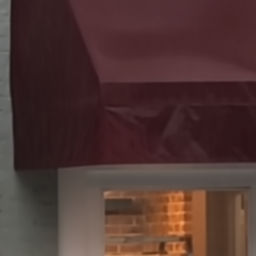}  
        \caption{Ground truth}  
    \end{subfigure}  
    \hfill  
    \begin{subfigure}[b]{0.3\linewidth}  
        \centering  
        \includegraphics[width=\linewidth,height=0.7\linewidth]{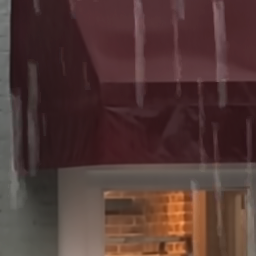}  
        \caption{Rainy (35.26 dB)}  
    \end{subfigure}  
    \hfill  
    \begin{subfigure}[b]{0.3\linewidth}  
        \centering  
        \includegraphics[width=\linewidth,height=0.7\linewidth]{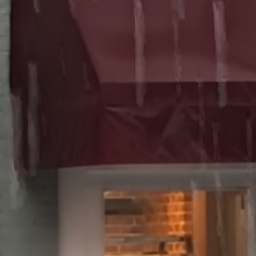}  
        \caption{DSC (32.36/0.915/0.050)}  
    \end{subfigure}  

    % 第二行子图  
    \vspace{0.5em}  
    \begin{subfigure}[b]{0.3\linewidth}  
        \centering  
        \includegraphics[width=\linewidth,height=0.7\linewidth]{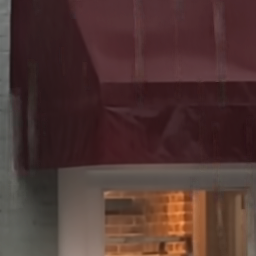}  
        \caption{RESCAN (38.65/0.962/0.024)}  
    \end{subfigure}  
    \hfill  
    \begin{subfigure}[b]{0.3\linewidth}  
        \centering  
        \includegraphics[width=\linewidth,height=0.7\linewidth]{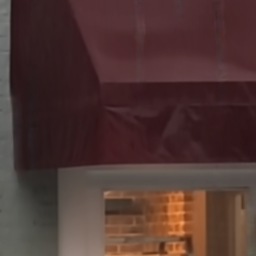}  
        \caption{MPRNet (45.62/\textbf{0.982}/0.011)}  
    \end{subfigure}  
    \hfill  
    \begin{subfigure}[b]{0.3\linewidth}  
        \centering  
        \includegraphics[width=\linewidth,height=0.7\linewidth]{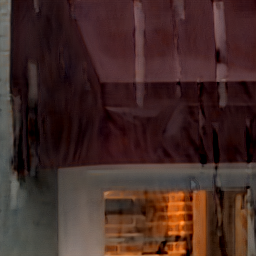}  
        \caption{SIRR (22.31/0.689/0.132)}  
    \end{subfigure}  

    % 第三行子图  
    \vspace{0.5em}  
    \begin{subfigure}[b]{0.3\linewidth}  
        \centering  
        \includegraphics[width=\linewidth,height=0.7\linewidth]{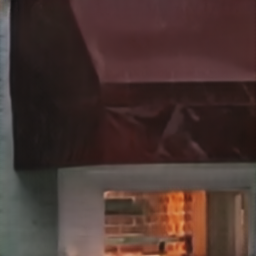}  
        \caption{CycleGAN (30.27/0.856/0.040)}  
    \end{subfigure}  
    \hfill  
    \begin{subfigure}[b]{0.3\linewidth}  
        \centering  
        \includegraphics[width=\linewidth,height=0.7\linewidth]{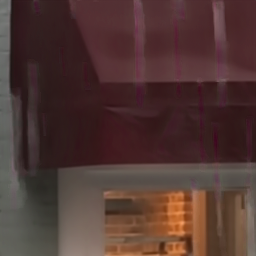}  
        \caption{DeCyGAN (34.58/0.918/0.053)}  
    \end{subfigure}  
    \hfill  
    \begin{subfigure}[b]{0.3\linewidth}  
        \centering  
        \includegraphics[width=\linewidth,height=0.7\linewidth]{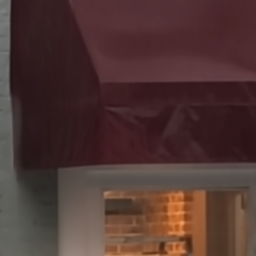}  
        \caption{SOT ($\ell_{1}$) (\textbf{45.72}/0.979/\textbf{0.008})}  
    \end{subfigure}  

    % 图片标题  
    \caption{Visual comparison on real-world image deraining. The PSNR/SSIM/LPIPS results are provided in the brackets.}  
    \label{figure5}  
\end{figure*}
Fig. \ref{figure5} compares the visual quality of 
the deraining methods on a typical real sample. 
It can be seen that, SOT can achieve a quality 
on par with the state-of-the-art supervised method MPRNet
to provide a visually plausible restoration, which
demonstrate the effectiveness of SOT on real data.
It should be noted that although CycleGAN can also achieve 
excellent rain removal, it introduces additional distortion such as color, 
resulting in larger distortion, e.g., with a PSNR more 
than 10 dB lower than that of MPRNet and SOT.

\subsection{Visual Samples of Synthetic Image Dehazing}
\label{SynDehsamples}

\begin{figure*}[!t]  
    \centering  
    % 第一行子图  
    \begin{subfigure}[b]{0.3\linewidth}  
        \centering  
        \includegraphics[width=\linewidth,height=0.7\linewidth]{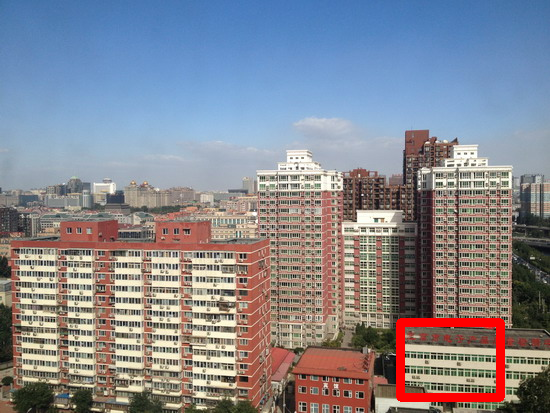}  
        \caption{Test image}  
    \end{subfigure}  
    \hfill  
    \begin{subfigure}[b]{0.3\linewidth}  
        \centering  
        \includegraphics[width=\linewidth,height=0.7\linewidth]{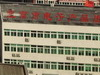}  
        \caption{Ground truth}  
    \end{subfigure}  
    \hfill  
    \begin{subfigure}[b]{0.3\linewidth}  
        \centering  
        \includegraphics[width=\linewidth,height=0.7\linewidth]{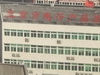}  
        \caption{Hazy (20.72 dB)}  
    \end{subfigure}  

    % 第二行子图  
    \vspace{0.5em}  
    \begin{subfigure}[b]{0.3\linewidth}  
        \centering  
        \includegraphics[width=\linewidth,height=0.7\linewidth]{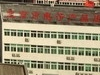}  
        \caption{DCP (18.23/0.851/0.067)}  
    \end{subfigure}  
    \hfill  
    \begin{subfigure}[b]{0.3\linewidth}  
        \centering  
        \includegraphics[width=\linewidth,height=0.7\linewidth]{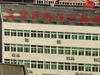}  
        \caption{Dehamer (\textbf{34.62}/\textbf{0.953}/0.017)}  
    \end{subfigure}  
    \hfill  
    \begin{subfigure}[b]{0.3\linewidth}  
        \centering  
        \includegraphics[width=\linewidth,height=0.7\linewidth]{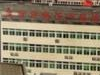}  
        \caption{GCANet (21.69/0.756/0.070)}  
    \end{subfigure}  

    % 第三行子图  
    \vspace{0.5em}  
    \begin{subfigure}[b]{0.3\linewidth}  
        \centering  
        \includegraphics[width=\linewidth,height=0.7\linewidth]{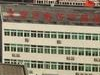}  
        \caption{FFANet (32.81/0.936/0.019)}  
    \end{subfigure}  
    \hfill  
    \begin{subfigure}[b]{0.3\linewidth}  
        \centering  
        \includegraphics[width=\linewidth,height=0.7\linewidth]{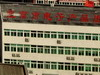}  
        \caption{D4 (21.98/0.864/0.047)}  
    \end{subfigure}  
    \hfill  
    \begin{subfigure}[b]{0.3\linewidth}  
        \centering  
        \includegraphics[width=\linewidth,height=0.7\linewidth]{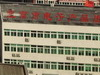}  
        \caption{SOT ($\ell_{1}$) (34.57/0.914/\textbf{0.016})}  
    \end{subfigure}  

    % 图片标题  
    \caption{Visual comparison on synthetic image dehazing. The PSNR/SSIM/LPIPS results are provided in the brackets. The images are enlarged for clarity.}  
    \label{figure6}  
\end{figure*}  
Fig. \ref{figure6} compares the visual quality of the methods. 
The restoration quality of the proposed method is even  
on par with Dehamer.

\subsection{Visual Samples of Real-world Image Dehazing}
\label{RealDehsamples}
\begin{figure*}[!t]  
    \centering  
    % 第一行子图  
    \begin{subfigure}[b]{0.3\linewidth}  
        \centering  
        \includegraphics[width=\linewidth]{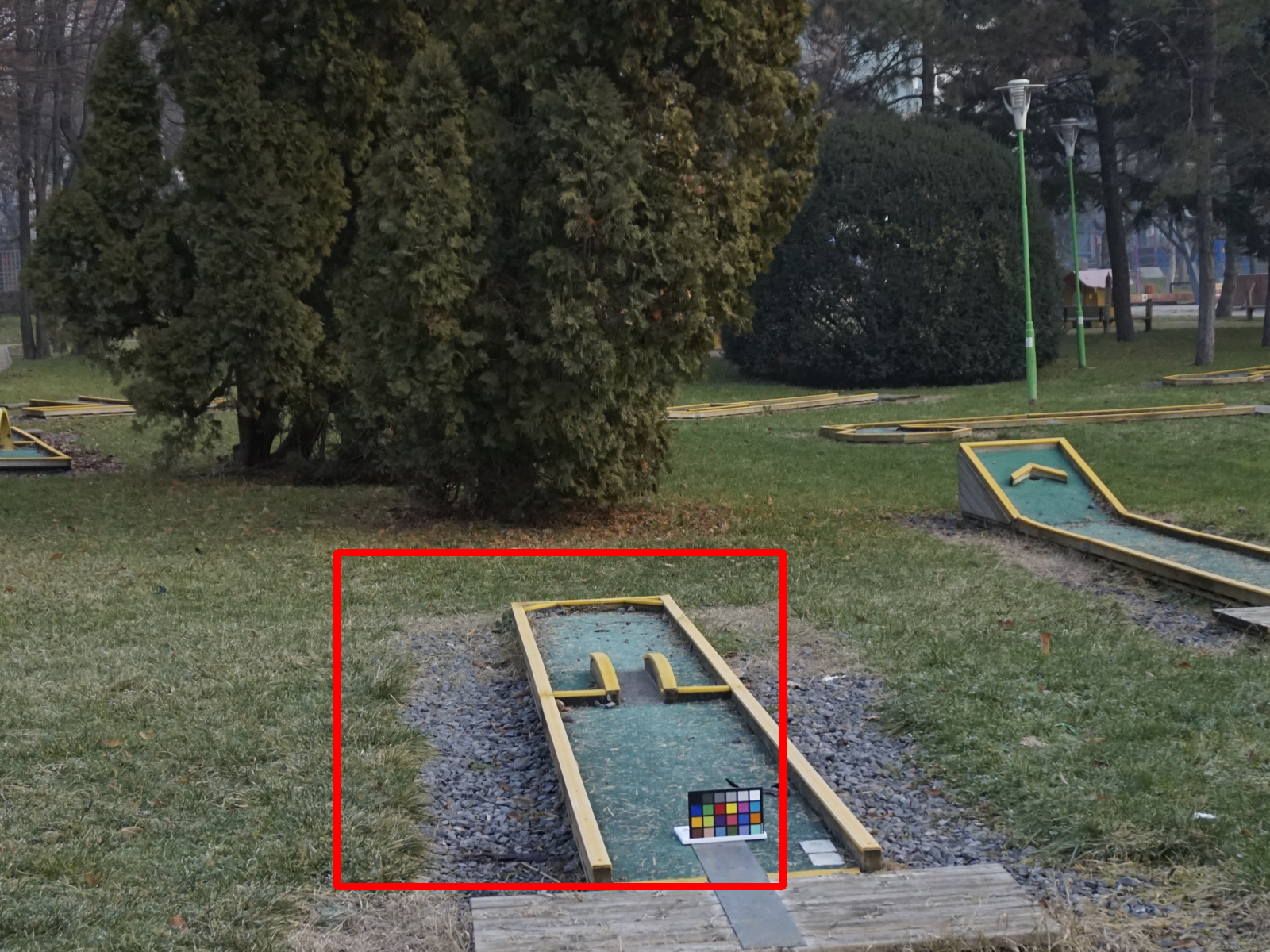}  
        \caption{Test image}  
    \end{subfigure}  
    \hfill  
    \begin{subfigure}[b]{0.3\linewidth}  
        \centering  
        \includegraphics[width=\linewidth]{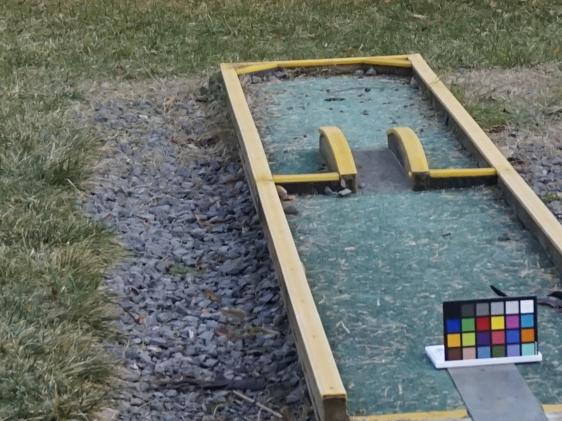}  
        \caption{Ground truth}  
    \end{subfigure}  
    \hfill  
    \begin{subfigure}[b]{0.3\linewidth}  
        \centering  
        \includegraphics[width=\linewidth]{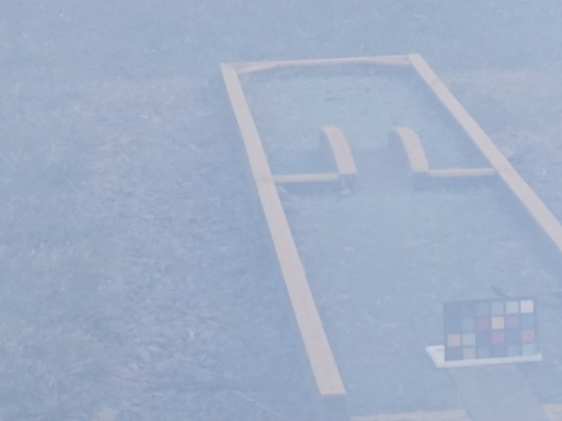}  
        \caption{Hazy (9.89 dB)}  
    \end{subfigure}  

    % 第二行子图  
    \vspace{0.5em}  
    \begin{subfigure}[b]{0.3\linewidth}  
        \centering  
        \includegraphics[width=\linewidth]{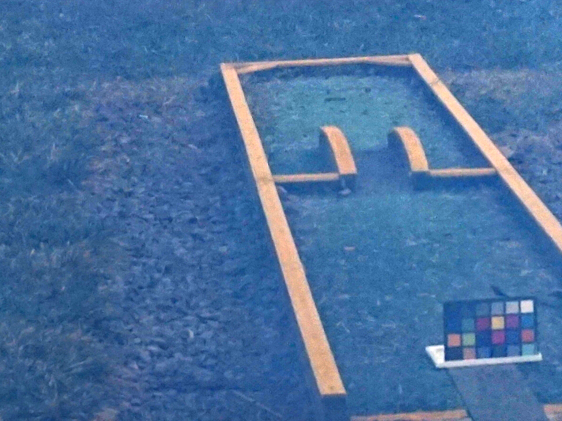}  
        \caption{DCP (10.35/0.407/0.342)}  
    \end{subfigure}  
    \hfill  
    \begin{subfigure}[b]{0.3\linewidth}  
        \centering  
        \includegraphics[width=\linewidth]{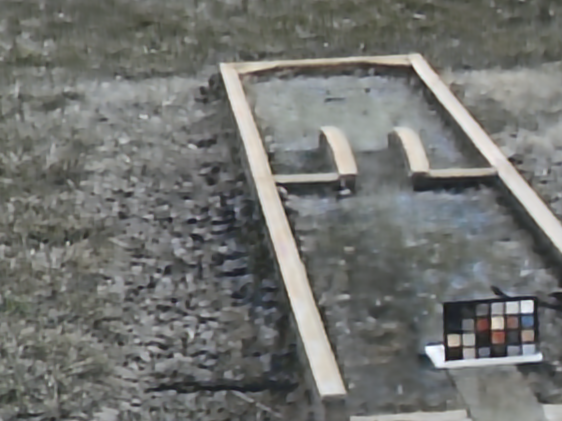}  
        \caption{Dehamer (\textbf{15.87}/\textbf{0.569}/0.155)}  
    \end{subfigure}  
    \hfill  
    \begin{subfigure}[b]{0.3\linewidth}  
        \centering  
        \includegraphics[width=\linewidth]{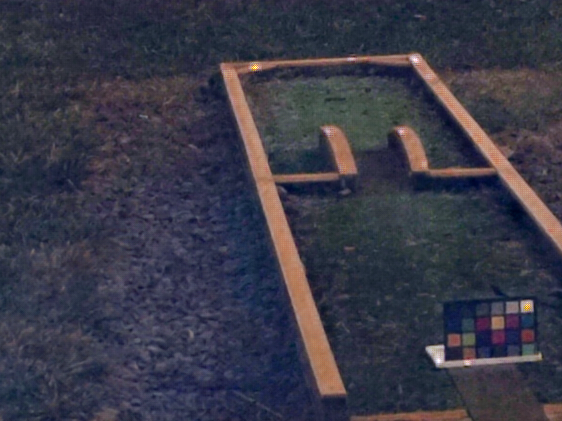}  
        \caption{GCANet (13.62/0.487/0.256)}  
    \end{subfigure}  

    % 第三行子图  
    \vspace{0.5em}  
    \begin{subfigure}[b]{0.3\linewidth}  
        \centering  
        \includegraphics[width=\linewidth]{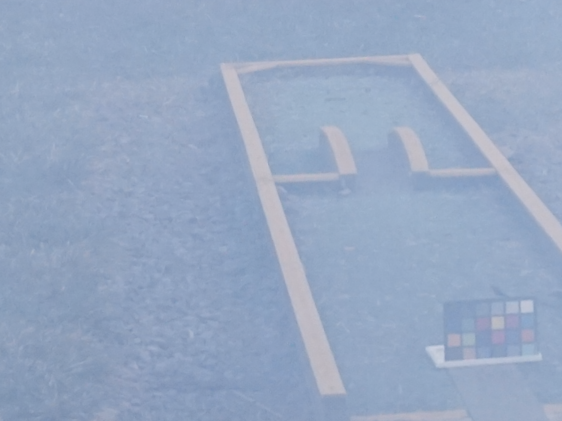}  
        \caption{FFANet (8.61/0.464/0.199)}  
    \end{subfigure}  
    \hfill  
    \begin{subfigure}[b]{0.3\linewidth}  
        \centering  
        \includegraphics[width=\linewidth]{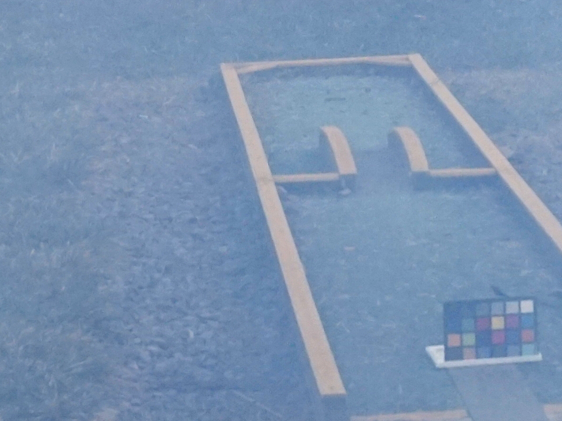}  
        \caption{D4 (9.77/0.471/0.217)}  
    \end{subfigure}  
    \hfill  
    \begin{subfigure}[b]{0.3\linewidth}  
        \centering  
        \includegraphics[width=\linewidth]{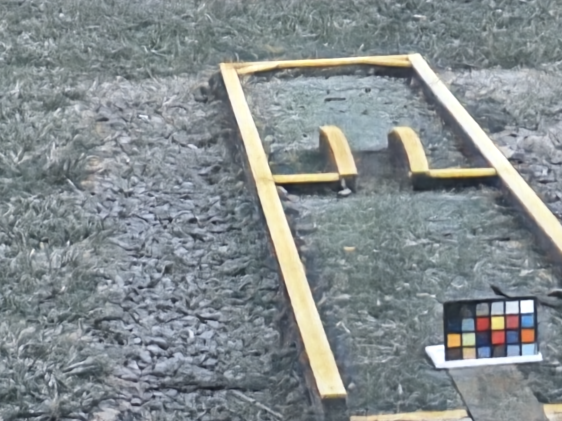}  
        \caption{SOT ($\ell_{1}$) (14.78/0.536/\textbf{0.146})}  
    \end{subfigure}  

    % 总标题  
    \caption{Visual comparison on a challenging real-world image dehazing task with severe haze.  
    The PSNR/SSIM/LPIPS results are provided in the brackets.  
    The images are enlarged for clarity.}  
    \label{figure7}  
\end{figure*}  
Fig. \ref{figure7} compares the visual quality of the methods. 
Noteworthily, the visual quality of SOT is distinctly better 
compared with the state-of-the-art transformer based supervised method Dehamer,
e.g. the color of the palette restored by SOT is much closer to the real scene.
This task is quite challenging as the observation (hazy images) 
is severely degraded with an average PSNR of only about 10 dB.
The results demonstrate the potential of the proposed method
on realistic difficult tasks to handle complex degradation.

\end{document}